\documentclass{article}

\usepackage[english]{babel}
\usepackage[letterpaper,top=2cm,bottom=2cm,left=3cm,right=3cm,marginparwidth=1.75cm]{geometry}

\usepackage{graphicx}
\usepackage{multicol,multirow}
\usepackage{amsmath,amssymb,amsfonts}
\usepackage{mathrsfs}
\usepackage{amsthm}
\usepackage{rotating}
\usepackage{appendix}
\usepackage{ifpdf}
\usepackage[T1]{fontenc}
\usepackage{newtxtext}
\usepackage{newtxmath}
\usepackage{textcomp}
\usepackage{xcolor}
\usepackage{lipsum}
\usepackage{verbatim}

\usepackage{float}
\usepackage{amssymb}
\usepackage{mathtools}
\usepackage{physics}
\usepackage{cancel}
\usepackage{tikz}
\usepackage{caption}
\usepackage{booktabs}
\usepackage{algorithm}
\usepackage{algcompatible}

\usepackage{url}
\usepackage[colorlinks,allcolors=blue]{hyperref}

\newcommand{\obs}       {y}
\newcommand{\nobs}      {p}
\newcommand{\ndelays}   {n}
\newcommand{\nterms}    {N_{\mathrm{terms}}}
\newcommand{\obsop}     {\mathbf{h}}
\newcommand{\state}     {u}
\newcommand{\nstate}    {m}
\newcommand{\stateop}   {\mathbf{f}}
\newcommand{\Lagr}     {\mathscr{L}}

\newcommand{\vecobs}    {\mathbf{\obs}}
\newcommand{\vecobstau}{\vecobs_{(\tau)}}
\newcommand{\vecobstaui}[1]   {\mathbf{\obs}_{(\tau_#1)}}
\newcommand{\vecstate}  {\mathbf{\state}}

\newcommand{\bx}        {\mathbf{x}}
\newcommand{\bz}        {\mathbf{z}}
\newcommand{\veca}      {\mathbf{a}}
\newcommand{\vecz}      {\mathbf{z}}
\newcommand{\vecpsi}    {\boldsymbol{\psi}}
\newcommand{\bpsi}      {\vecpsi}
\newcommand{\bvarphi}   {\boldsymbol{\varphi}}
\newcommand{\btheta}    {{\boldsymbol{\theta}}}
\newcommand{\blambda}   {\boldsymbol{\lambda}}
\newcommand{\embedd}    {\mathfrak{\obs}}
\newcommand{\lrate}     {\eta}

\newcommand{\Liouvilleop} {\mathcal{L}}
\newcommand{\bnabla} {\boldsymbol{\nabla}}

\newcommand{\ddroit}    {\mathrm{d}}

\newcommand{\ie}        {\textit{i.e.}}
\newcommand{\eg}        {\textit{e.g.}}
\newcommand{\bzero}     {\boldsymbol{0}}

\DeclareMathOperator*{\minLM}	{{\rm min}}

\newcommand{\noise}{\boldsymbol{\xi}}

\theoremstyle{plain}
\newtheorem{theorem}{Theorem}[section]
\newtheorem{proposition}[theorem]{Proposition}

\theoremstyle{definition}

\theoremstyle{remark}

\numberwithin{equation}{section}

\title{Neural delay differential equations: learning non-Markovian closures for partially known dynamical systems}

\author{Thibault Monsel$^1$, Onofrio Semeraro$^2$, Guillaume Charpiat$^3$, Lionel Mathelin$^2$
\\
\\
$^1$ LISN, Universit\'e Paris-Saclay, Orsay, 94100 France
\\
$^2$ CNRS, LISN, Universit\'e Paris-Saclay, Orsay, 94100 France
\\
$^3$ INRIA, LISN, Universit\'e Paris-Saclay, Gif-sur-l'Yvette, 91190 France
}

\date{}

\begin{document}
\maketitle


\begin{abstract}
Recent advances in learning dynamical systems from data have shown significant promise. However, many existing methods assume access to the full state of the system -- an assumption that is rarely satisfied in practice, where systems are typically monitored through a limited number of sensors, leading to partial observability. 
To address this challenge, we draw inspiration from the Mori-Zwanzig formalism, which provides a theoretical connection between hidden variables and memory terms. Motivated by this perspective, we introduce a constant-lag Neural Delay Differential Equations (NDDEs) framework, providing a continuous-time approach for learning non-Markovian dynamics directly from data. These memory effects are captured using a finite set of time delays, which are identified via the adjoint method. 
We validate the proposed approach on a range of datasets, including synthetic systems, chaotic dynamics, and experimental measurements, such as the Kuramoto-Sivashinsky equation and cavity-flow experiments. Results demonstrate that NDDEs compare favourably with existing approaches for partially observed systems, including long short-term memory (LSTM) networks and augmented neural ordinary differential equations (ANODEs). Overall, NDDEs offer a principled and data-efficient framework for modelling non-Markovian dynamics under partial observability. An open-source implementation accompanies this article.
\end{abstract}


\section{Introduction}

Describing the dynamics of a system is instrumental in many domains such as biology \cite{Roussel1996,Epstein1990DifferentialDE}, climate research \cite{nino,nino2}, or finance \cite{Achdou2012PartialDE}. Indeed, an accurate description of the evolution of a system not only improves knowledge and understanding but also allows forecasting, a critical requirement in many situations where decisions must be taken based on predictions, or when devising a suitable sequence of actions to achieve some goal requires a good knowledge of the effect of these actions on the system under consideration. The system is typically supposed to be governed by a model of the form
\begin{equation}
\dv{\vecstate(t)}{t} = \stateop(\vecstate(t)), \qquad \vecstate(t_0) = \vecstate_0,
\label{eq:intro_model}
\end{equation}
with $\vecstate(t)$: $\mathbb{R} \to \mathbb{R}^{\nstate}$ the time-dependent solution vector representing the state of the system approximated in a suitable finite-dimensional representation basis and $\stateop$: $\mathbb{R} \times \mathbb{R}^{\nstate} \to \mathbb{R}^{\nstate}$ the vector field. A reliable model may, however, not be available and one then has to infer it from monitoring the system, here in the form of $L$ distinct trajectories. This model-free approach solely relies on the history of observational data to learn a model for predicting their dynamics. One can use the dataset to fit the ODE dynamics model Eq.~\eqref{eq:intro_model}, and many previous efforts have been reported toward this aim. For instance, Neural Ordinary Differential Equations (NODEs), introduced in \cite{chen2018neural}, follow this exact formulation. NODEs gave rise to the class of continuous-depth models and can be viewed as a continuous extension of Residual Networks \cite{resnet}. An immediate extension of NODEs, referred to as Augmented NODEs \cite{dupont2019augmented}, explores the existence of certain functions that NODEs are unable to represent. It tackles the expressivity limitation of NODEs, {where expressivity is here informally accepted as the richness of the class of functional representations}, by expanding the dimension of the solution space through the incorporation of additional variables to learn more complex functions using simpler flows.

\medskip
However, the state $\vecstate$ is not necessarily available and, in most situations of practical interest, the sole source of information about a system is via a small set of sensors providing a few observations. In addition, the system may not be Markovian and may adopt a behavior which depends not only on its current state but also on its history. 

Relying on the past to compensate for the lack of information from the current state is a common approach in partially observed systems where the available observations are not sufficient statistics to predict the future evolution deterministically. One then reverts to predicting an uncertain future, in the form of a probability distribution as is done with Kalman filters, or to accounting for past measurements to narrow down the possible futures consistent with the observations to a unique one. The problem hence formulates as time series prediction. State-of-the-art methods for time series prediction involve autoregressive models like ARMAX \cite{guidorzi2003multivariable} and related algorithms; recurrent neural networks such as LSTM, whose distinctive feature is their ability to incorporate a "memory" as a latent variable \cite{rnn_intro, rnn_intro2, Hochreiter1995LONGST}; echo state networks and reservoir computing \cite{jaeger2004harnessing,maass2002real}, often regarded as an alternative to RNNs for their efficient training and strong performance in capturing long-term statistics when full state dynamics are accessible \cite{vlachas2020backpropagation}. Other techniques encompass Latent ODE, which combines NODEs and RNNs together \cite{latent_ode} -- a variational autoencoder model using an ODE-RNN encoder and ODE decoder architecture to construct a continuous-time model with a latent state defined at all times -- the already mentioned Augmented Neural ODEs \cite{dupont2019augmented}, or the recently introduced Transformers \cite{Transformers_17}. While often effective, these techniques lack expressivity (ARMAX, limited in its ability to account for the complex dynamics) or interpretability (recurrent networks, Transformers), in the sense that their resulting architecture cannot be clearly justified by, or related to, specific considerations on the system, in contrast with time-delays which can be linked with physical feed-back mechanisms and time-constants for instance. Further, the discrete nature of some of these approaches collides with the continuous formulation of the problem considered in this work, where the focus is on learning a model for a dynamical system from observations, possibly complementing an already available imperfect or coarse model, then formulated as a closure problem.

\medskip
A classical framework that formalizes these ideas is the Mori-Zwanzig (MZ) projection formalism, originally developed in statistical physics to derive macroscopic equations for systems with many degrees of freedom. The MZ formalism decomposes the full dynamics into three terms: a Markovian contribution from the directly resolved variables, a memory term that encodes the cumulative influence of the unresolved modes, and a fluctuating noise term. The resulting Generalized Langevin Equation (GLE) provides a mathematically exact description of the reduced system, where the effect of missing physics appears as a convolution integral over the past. In practice, however, this integro-differential form is rarely tractable, motivating the search for parsimonious representations of the memory kernel. Within fluid mechanics, the MZ formalism has increasingly been considered as a theoretical foundation for closure modeling and reduced-order modeling. In \cite{Parish2016dynamic, Parish2017unified, parish2017non}, it is shown that finite-memory approximations of the MZ kernel can yield dynamic subgrid-scale models, connecting the MZ projection to large-eddy simulation (LES) closures. These studies also extended to variational multiscale formulations, emphasizing that MZ closures naturally align with hierarchical scale separation and energy transfer mechanisms. \cite{LinTian2021} developed algorithms to estimate the Markov, memory, and noise terms directly from simulation data, demonstrating that such operator learning can capture long-term correlations absent in purely Markovian models. In the context of neural models, a recent work by \cite{gupta2025mori} explored MZ-based closures in latent spaces obtained from nonlinear autoencoders, showing that embedding MZ dynamics into reduced coordinates can enhance stability and interpretability. The interested reader can refer to the recent overview by \cite{sanderse2024scientific}.

\medskip
Another approach is to explicitly incorporate non-Markovianity into the formulation by including the historical past state in Equation~\eqref{eq:intro_model}. This leads to the domain of Neural Delay Differential Equations (NDDEs), which constitutes another subset of models falling under the umbrella of continuous-depth models, alongside NODEs. A generic delay differential equation (DDE) is described by:
\begin{equation}
\begin{aligned}
 \dv{\vecstate}{t} & = \stateop\left(\vecstate(t), \vecstate(\alpha_1(t)), \dots, \vecstate(\alpha_\ndelays(t))\right), \qquad \vecstate(t \leq 0) = \vecpsi(t), \\
 \alpha_i(t) & := t - \tau_i(t, \vecstate(t)), \qquad i \in \{1, 2, \ldots, \ndelays\},
\end{aligned}
\label{eqn:dde}
\end{equation}
where $\vecpsi(t)$: $\mathbb{R}^- \to \mathbb{R}^\nstate$ is the history function, $\tau_i$: $\mathbb{R} \times \mathbb{R}^\nstate \to \mathbb{R}^+$ is a delay function, and $\stateop$ is the vector field. The history function $\vecpsi$ serves as the initial condition for DDEs, analogous to $\vecstate_0$ in ODEs. The modeling capabilities of NDDEs vary based on the chosen type of delay. Inherently, NDDEs incorporate and leverage information from preceding time points, effectively converting the delay term into a dynamic memory mechanism. Initially proposed by \cite{zhu2021neural} to learn NDDEs with a single constant delay, subsequent work by \cite{zhu2023neural} and \cite{schlaginhaufen2021learning} explored piecewise constant delays and developed a stabilizing loss for NDDEs, respectively. Additionally, \cite{oprea2023learning} focused on learning a single delay within a small network (fewer than 10 parameters). In \cite{monsel2024time}, we revisit the Neural DDE by introducing Neural State-Dependent DDE (SDDDE), a general and flexible framework that can model multiple and state- and time-dependent delays.

\medskip
Coming back to the MZ formalism, \cite{GuptaLermusiaux2021} demonstrated that neural delay differential equations (NDDEs) could represent non-Markovian effects in a tractable way, motivating the idea of combining neural networks with delay-based memory embeddings within the MZ framework. In a similar spirit, \cite{Menier_JCP} introduced an exponentially decaying memory, allowing for data-efficient training and interpretability of the decay rate in terms of the time scales of the underlying system. In this work, we extend our previous contributions \cite{monsel2024time} by embedding NDDEs in the general framework of the MZ and Takens formalisms along similar lines as \cite{GuptaLermusiaux2021}.
{Specific to the present work is that the multiple delays are here \emph{learned} rather than being fixed a priori. To this aim, an adjoint formulation is derived and allows efficient learning. The impact of relevant delays onto the information extracted from the past is illustrated in Section~\ref{App_MI}.} We explore the possibility of learning the values of the delays by deriving a practical training procedure with learnable delays using adjoint differentiation. The adjoint procedure makes NDDEs computationally efficient for large datasets; we illustrate the approach on both synthetic and experimental systems, including chaotic dynamics, partially observed fluid flows, and reduced-order closure models, highlighting how the learned delays may correspond to physical time scales.

The remainder of the paper is organized as follows. Section \ref{sec:MZsection} revisits the MZ formalism and discusses how time delays can be used for representing the memory kernel. Section \ref{sec:ndde_ld} details the proposed NDDE formulation, presents the adjoint training method with learnable delays, and outlines its implementation. We report numerical and experimental validations in Section \ref{results}. In Section \ref{sec:dderom}, NDDEs are used as closure models within a reduced-order framework. Section \ref{sec:conclusion} summarizes the findings, discusses limitations, and outlines future research directions.


\section{Modeling Partially Observed Dynamical Systems}
\label{sec:MZsection}

Let us consider the nonlinear system stated in Eq.~\eqref{eq:intro_model}:
\begin{equation}
\begin{aligned}
\dv{\vecstate(t)}{t} = \stateop(\vecstate(t)), \qquad \vecstate(t_0)=\vecstate_0,
\end{aligned}
\label{th:mztheorem}
\end{equation}
with $\vecstate(t)$ evolving on a smooth manifold $\mathcal{S} \subset \mathbb{R}^\nstate$.
We assume the system is only observed through a small set of sensor functions belonging to a Hilbert subspace of $L^2(\mathcal{S},\mathbb{R})$ providing measurements $\left\{\obs_i\right\}_{i=1}^{\nobs}$, $\obs_i(\vecstate(t)): \mathcal{S} \to \mathbb{R}$ and let $\vecobs(\vecstate(t)) = \left(\obs_1, \ldots, \obs_{\nobs}\right)$. The dynamics of these observables can be derived from the model of the system:
\begin{equation}
    \frac{\partial \vecobs(\vecstate(t))}{\partial t} = \bnabla_\vecstate \vecobs \:\: \stateop(\vecstate(t)) \equiv \Liouvilleop \,\vecobs(\vecstate(t)),
\end{equation}
where $\Liouvilleop$ is linear and termed a Liouville operator. Defining the Liouville operator at $t=t_0$ yields $\displaystyle \Liouvilleop \, \vecobs = \bnabla_{\vecstate_0} \vecobs(\vecstate_0) \:\: \stateop(\vecstate_0)$ and the observables then express as $\vecobs(\vecstate(t)) = e^{(t-t_0) \Liouvilleop} \, \vecobs(\vecstate_0)$.

Because the measurements are limited in number, the span $\mathcal{G} := \mathrm{span}\left(\left\{\obs_i\right\}_i: \mathcal{S} \to \mathbb{R}\right)$ is finite, low-dimensional, and not invariant under the time-evolution operator governing the dynamics of $\vecobs(\vecstate(t > t_0))$. The dynamics of $\vecobs$ is then not closed in $\mathcal{G}$, reflecting the fact that some information leaks in an orthogonal subspace $\mathcal{G}^\perp$. To recover the missing information, it is necessary to describe how to express the dynamics of $\vecobs$ not just as a function of itself but also accounting for its history. The rationale is that the past values of $\vecobs$ are influenced by unobserved quantities, or unresolved observables, of the system not contained in $\mathcal{G}$. Carefully processing the history of these observables then allows to account for the impact of the unresolved observables onto the observables $\vecobs$. 

\subsection{The Mori-Zwanzig formalism}

Different approaches leverage this past information. 
The Mori-Zwanzig (MZ) formalism, rooted in statistical physics, provides a framework to derive an evolution equation for a set of variables, such as macroscopic observables, associated with a high-dimensional dynamical system \cite{mori_intro, zwanzig_intro1, zwanzig_intro2}. This framework is instrumental, for instance in situations where the full state $\vecstate(t)$ is unavailable and where one can only access low dimensional observations. Similarly, the MZ formalism is also relevant for addressing dimension reduction problems \cite{zhu2019mori} where the reduced state is a set of variables whose dynamics are affected by the projected-out and non-observed variables or state components.

In the Mori-Zwanzig framework, the span of $\mathcal{G}$ of the measurements is considered as resulting from a projection through a projector $\mathcal{P}$ of the space of $L^2(\mathcal{S},\mathbb{R})$ functions. Similarly, the unobserved functions span a subspace resulting from a projection $\mathcal{Q}$, with $\mathcal{Q} = \mathcal{I} - \mathcal{P}$. The dynamics of $\vecobs$ can then be expressed as

\begin{equation}
    \frac{\partial}{\partial t} \vecobs(\vecstate(t))  = \frac{\partial}{\partial t} e^{(t-t_0) \Liouvilleop } \vecobs(\vecstate_0) = e^{(t-t_0) \Liouvilleop} \left(\mathcal{P}+\mathcal{Q}\right) \Liouvilleop \, \vecobs(\vecstate_0).
\end{equation}
The influence of past unresolved dynamics onto the current observations can be explicitly expressed using the Dyson identity which states:
\begin{equation}
e^{(t-t_0) \Liouvilleop } = \int_{0}^{t-t_0}{e^{(t-t_0-s) \Liouvilleop} \mathcal{P} \Liouvilleop \, e^{s \mathcal{Q} \Liouvilleop} \ddroit s} \, + \, e^{(t-t_0) \mathcal{Q} \Liouvilleop},
\end{equation}
so that one can write the dynamics of the current observables $\vecobs(\vecstate(t))$ as:
\begin{align}
   \frac{\partial}{\partial t} \vecobs(\vecstate(t)) = \mathcal{P} \Liouvilleop \, \vecobs(\vecstate(t)) + \int_{0}^{t-t_0}{\mathcal{P} \Liouvilleop \, e^{s \mathcal{Q} \Liouvilleop} \mathcal{Q} \Liouvilleop \, \vecobs(\vecstate(t-s)) \ddroit s} \, + \, e^{(t-t_0) \mathcal{Q} \Liouvilleop} \mathcal{Q} \Liouvilleop \, \vecobs(\vecstate_0).
   \label{eq_preGLE}
\end{align}

Rewriting Eq.~\eqref{eq_preGLE} in a more compact form, and denoting $\vecobs(\vecstate(t)) = \vecobs(t)$, the dynamics of the vector of observables $\vecobs=\left(\obs_1, \dots, \obs_\nobs\right)$ are shown to follow an Integro-Differential Equation (IDE):
\begin{equation}
\begin{aligned}
\frac{\partial \vecobs(t)}{\partial t} = M \, \vecobs(t) + \int_{0}^{t-t_0}{K(\vecobs(t-s),s) \, \ddroit s} + F(t,\vecstate_0),
\label{eqn:mz_no_noise}
\end{aligned}
\end{equation} 
with $M$ a Markovian operator and $K$ an operator applied to past observables and integrated over the whole time span since the initial condition. The so-called ``noise'' term $F$ accounts for the impact of the unobserved variables at the initial condition. It can be eliminated from the equations predicting future observations by applying the projector $\mathcal{P}$ and restricting their dynamics into the linear span of $\mathcal{P} \vecobs$.
Learning the dynamics of partially observed or known systems hence boils down to estimating each term of the differential equation above, sometimes referred to as the Generalized Langevin Equation (GLE).

\medskip
Accounting for the past essentially allows to isolate the dynamics of the observables. The Mori-Zwanzig framework is general and applies widely, allowing to describe the dynamics of the observables with a non-Markovian model. It similarly provides a principled closure for coarse models, hence missing information, which can be effectively complemented with a history-based term.

\begin{figure}[t]
\centering
\includegraphics[scale=0.2, trim=1.5cm 4cm 4cm 1.5cm]{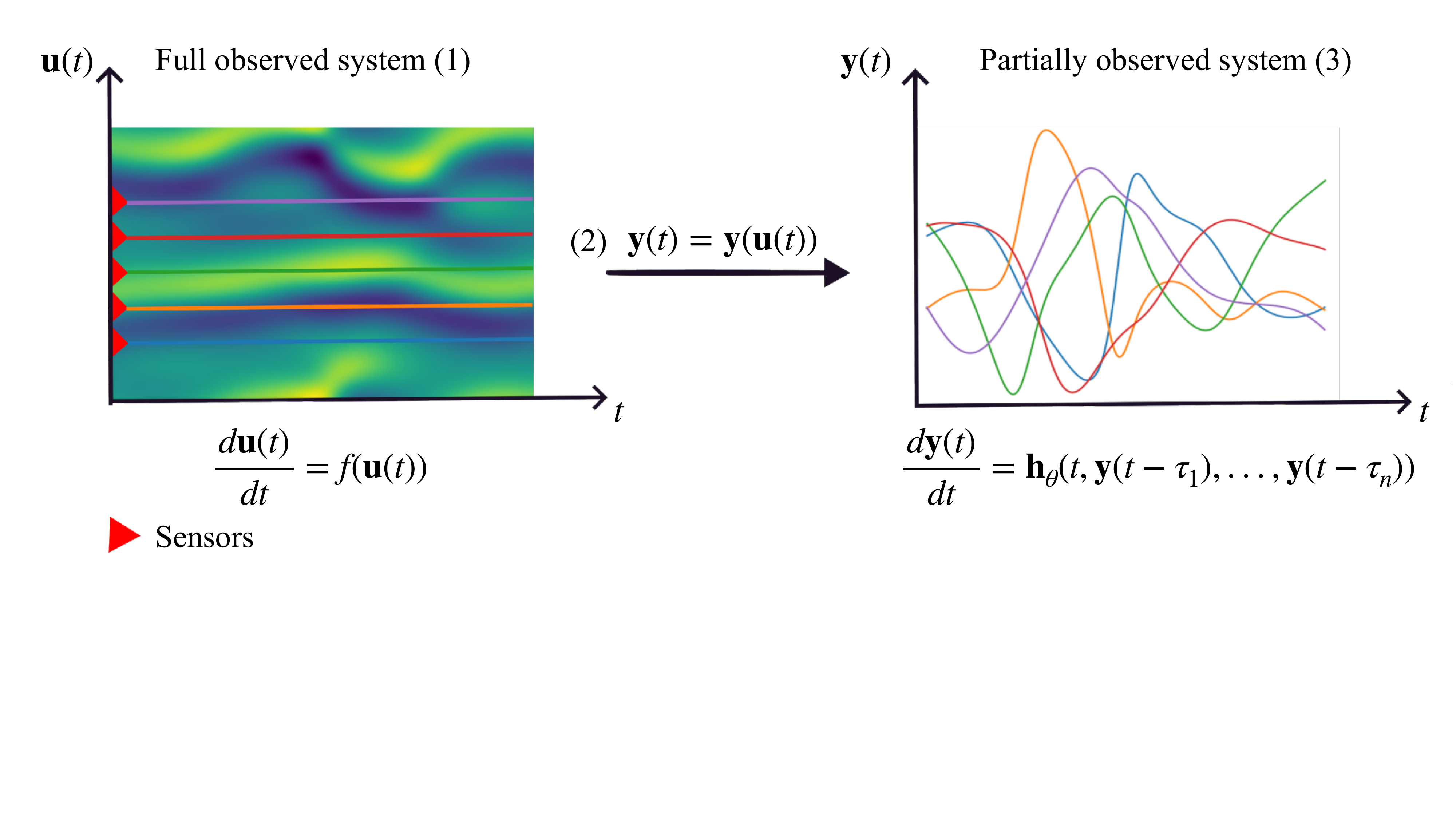}
\vspace{-1.25cm}
 \caption{Measuring the fully observed state of a system (left (1)) is often impossible due to  its high dimensionality. Ultimately the user only has at its disposable sparse observations of the full state that can be seen as low dimensional observable vector $\vecobs$ (middle (2)). The MZ equation DDE approximation (Proposition~\ref{prop:dde}) can then used to model partially observed systems (right (3)).}
\label{fig:mz_expose}
\end{figure}


\subsection{Approximations of Integro-Differential Equations (IDE)}

The statement in Eq.~\eqref{eqn:mz_no_noise} outlines the structure of the dynamics. However, the operators involved in this dynamical equation are usually poorly known, if any, in particular the integral term. We now discuss several approaches from the literature to estimate these terms.

An approach to approximate the integral consists in studying particular asymptotic regimes. Among the many different models proposed in the literature \cite{stinis2003stochastic,chorin2005problem,stinis2006higher}, one of the most popular consists in approximating the integral under assumptions of very short memory or very long memory regimes. For example, the \textit{t-model} \cite{chorin2002optimal}, also commonly called \textit{slowly decaying memory approximation}, leads to Markovian equations with time-dependent coefficients, which can subsequently be modeled using a NODE. However these remain asymptotic approximations, and cannot be extended to intermediate-range memory, in general.
Another approach consists in performing  Monte Carlo integration \cite{robert1999monte}. This approach has been extended to a neural network-based formulation (Neural IDE)~\cite{zappala2022neural}, where the memory integrand is decomposed as a product $K_1(t,s) \, K_2(\vecobs(s))$. However the number of function evaluations required to accurately integrate Equation~\eqref{eqn:mz_no_noise} can be large, making the process computationally intensive (cf. Appendix~\ref{app:idevsdde}). In a similar spirit, under the assumption of short memory, one can restrict the integral to a short past and discretize it in time, leading to an equation of the form \cite{gallage2017approximation}:
\begin{equation}
\begin{aligned}
\frac{\partial \vecobs(t)}{\partial t} \approx M(\vecobs(t)) + \frac{1}{\ndelays} \sum_{i=1}^{\ndelays}{K(\vecobs(t-\tau_i), \tau_i)},
\label{eqn:mz_discr}
\end{aligned}
\end{equation} 
using $\ndelays$ delays $\left\{\tau_i\right\}_i$ uniformly spaced over a past horizon instead of sampling them randomly. Such approximations can be improved using high-order discretization schemes, yet, as for Neural IDE, they require an unaffordable number of delays if the integrand varies quickly or if the time interval is too large.


\subsection{Exact representation with Neural DDE}

While Equation~\eqref{eqn:mz_discr} only provides an approximation of the true dynamics and requires many delays, we show that using a more complex function of a small number of delays, it is actually possible to represent the dynamics \emph{exactly}. For this, let us consider diffeomorphic dynamical systems, that is, ODEs whose flow is invertible (in the full state space) and smooth in both time directions.
\begin{proposition}[\textbf{Exact representation with delays}]\label{prop:dde}

For any smooth dynamical system ($C^2$ is enough), and differentiable observables $\vecobs$, there exists almost surely an operator $M$ of the current observables, a finite number $\ndelays$ of delays $\tau_1, ..., \tau_{\ndelays} > 0$ and a function $\obsop$ such that the observables \emph{exactly} follow the dynamics:
\begin{equation}
\begin{aligned}
\frac{\partial \vecobs(t)}{\partial t} = M(\vecobs(t)) + \obsop\left(t, \vecobs(t), \vecobs(t-\tau_1), \vecobs(t-\tau_2), \ldots, \vecobs(t-\tau_\ndelays)\right).\label{eqn:neuraldde}
\end{aligned}
\end{equation} 
\end{proposition}
The derivation of the above, deferred to Appendix~\ref{app:takens}, is based on the application of Takens' theorem (hence the diffeomorphism requirement), which also provides a bound on the required number of delays: at least twice the intrinsic dimension of the manifold $\mathcal{S}$ in which the full state $\vecstate$ lives. Note that this evolution equation is exact: approximations may arise from the optimization or the expressivity of the neural networks estimating $M$ and $\obsop$, but not from the number of delays, provided they reach the Takens' lower bound. This is in contrast with the discretization of the IDE integral as in Equation~\eqref{eqn:mz_discr}, which becomes asymptotically precise only when the number of delays is large compared to the complexity of the integrand.


\newpage

\section{Neural Delay Differential Equations with Learnable Delays}\label{sec:ndde_ld}
The proposition~\ref{prop:dde} above motivates our approach. We consider learning the dynamics of the observables by accounting for their past in the form of a set of time-continuous coupled delayed differential equations (DDEs) involving a number of past observations $\left\{\vecobs(t-\tau_1), \ldots, \vecobs(t-\tau_{\ndelays})\right\}$. The vector field is approximated by a neural network and the resulting model is solved by NDDE, a Neural Differential Equation solver dedicated to delayed equations, \cite{monsel2024time}. The Mori-Zwanzig framework provides a rigorous rationale, grounded in statistical physics, on how the dynamics of the observables is related to their past, involving a continuous integral over a past horizon. In contrast, the Takens' theorem provides a geometrical view of the dynamics of a set of observables as a function of a sufficiently large number of delayed versions of themselves. These two frameworks motivate our approach of modeling the dynamics of the observables from a \emph{finite} set of past observations via a learnable function. In addition, while the Takens' theorem is not constructive when it comes to the value of the time lag, we here learn every individual delay $\left\{\tau_i\right\}$, $1 \le i \le \ndelays$. The minimum number of delays $\ndelays$ is determined from the $D_2$ dimension of the underlying system, estimated from the dataset of observations.

\subsection{Learning the delays}\label{Sec_learning_delays}
A constant lag NDDE is part of the larger family of continuous depth models that emerged with NODE \cite{chen2018neural}. It is defined by:
\begin{equation}
\pdv{\vecobs(t)}{t} = \obsop_{\btheta}(t, \vecobs(t), \vecobs(t-\tau_1), \dots, \vecobs(t-\tau_{\ndelays})), \qquad \vecobs(t \leq 0) = \bpsi(t), \\
\label{eqn:constantndde}
\end{equation}
with $\bpsi$: $\mathbb{R} \to \mathbb{R}^\nstate$ the history function, $\tau_i \in \mathbb{R}^{+}$ a delay constant and $\obsop_{\btheta}: [0, T] \times \mathbb{R}^\nstate \times \dots \times \mathbb{R}^\nstate \to \mathbb{R}^\nstate$ a neural network. 

There are two possible ways of training continuous-depth models:~discretize-then-optimize or optimize-then-discretize \cite{kidger2022neural}. In the former, the numerical simulation library's inherent auto-differentiation capabilities are leveraged. In the latter, the adjoint dynamics are employed to compute the gradient's loss. The following proposition \ref{th:adjoint_dde} provides the adjoint method for constant lag NDDEs.

\begin{proposition}[]\label{th:adjoint_dde}
Let us consider the Neural DDE model below, where $\tau$ is a learnable vector parameterized by some components of $\btheta$:
\begin{equation}
\frac{\partial \vecobs(t)}{\partial t}  = \obsop_\btheta(t, \vecobs(t), \vecobs(t-\tau)), \qquad \vecobs(t\leq 0) = \bpsi(t), \qquad 0 \le t \le T,
\label{eqn:single_delay_dde}
\end{equation}
and the following loss function:
\begin{equation*}
	L(\vecobs) = \int_0^T{l(\vecobs(t))\, \ddroit t}.
\end{equation*}
The gradient of the loss with respect to the parameters is given by: 
\begin{align}
    \begin{split}
        \pdv{L}{\btheta} = & - \int_0^T{\blambda(t) \left(\pdv{\obsop_\btheta(t, \vecobs(t), \vecobs(t - \tau))}{\btheta} - \pdv{\obsop_\btheta(t, \vecobs(t), \vecobstau(t))}{\vecobstau} \pdv{\vecobs(t - \tau)}{\btheta} \right) \ddroit t},
    \label{eq:grad_wtr_param}
    \end{split}
\end{align}
where $\vecobstau(t) \equiv \vecobs(t-\tau)$ is the third set of variables $\obsop_\btheta(t,\vecobs,\vecobstau)$ depends on. The adjoint dynamics $\blambda(t)$ are given by another DDE: 
\begin{align}
    \begin{split}
    & \dv{\blambda(t)}{t} = \pdv{l(\vecobs(t))}{\vecobs} - \blambda(t) \pdv{\obsop_\btheta(t,\vecobs(t),\vecobstau(t))}{\vecobs} - \blambda(t+\tau) \pdv{\obsop_\btheta(t+\tau,\vecobs(t+\tau),\vecobstau(t+\tau))}{\vecobstau},\\
	&\blambda(t\geq T) = \bzero.
\label{eq:adjoint_dyn}
\end{split}
\end{align}
\end{proposition}
The derivation of the Proposition~\ref{th:adjoint_dde} is in Appendix~\ref{app:proof_adjoint}.  Algorithm~\ref{alg:algo_multiple_ndde} below outlines the training procedure for a Neural DDE with multiple learnable constant delays, defined for convenience as $\vecobstaui{i}(t) = \vecobs(t-\tau_i)$. 

The developments are packaged in a user-friendly API, developing a numerically robust DDE solver, and implementing the adjoint method in the \texttt{torchdde} package. These advancements allow a seamless integration of DDEs for future users, enhancing reproducibility. Computational details and benchmarks are reported in Appendix~\ref{torchddebench}.


\begin{algorithm}[H]
\caption{Training a Neural DDE with learnable delays with the adjoint method.}
\label{alg:algo_multiple_ndde}
\begin{algorithmic}[1]
\REQUIRE Dataset of one trajectory $\mathcal{D} = \{(t_0, \vecobs_0), \dots, (t_{T}, \vecobs_{T}) \}$.
\REQUIRE Initialized model $\obsop_\btheta$.
\REQUIRE Initialized positive delays $\tau_1, \dots, \tau_{\ndelays}$, handled as additional entries in the parameters vector $\btheta$.
\FOR{$N_{\rm epochs}$}
    \STATE Set $\tau_{\rm max} = \max\left\{\tau_1, \dots, \tau_{\ndelays}\right\}$
    \STATE Create history function interpolation $\psi$ with data from $\mathcal{D}$ such that $t \leq t_0+\tau_{\rm max}$.
    \STATE Solve DDE dynamics: \\
    $\begin{cases}
      \displaystyle \frac{\ddroit \vecobs(t)}{\ddroit t} = \obsop_{\btheta}(t, \vecobs(t), \vecobstaui{1}(t), \dots, \vecobstaui{\ndelays}(t)) \\
      \vecobs(t \leq t_0+\tau_{\rm max}) = \bpsi(t)
     \end{cases}\,$ 
    \STATE Compute loss $\displaystyle L(\vecobs) = \int_{t_0+\tau_{\rm max}}^{t_T} l(\vecobs(t))\, \ddroit t$ 
    \STATE Solve Adjoint dynamics: \\ $\begin{cases} \displaystyle
      \dot{\blambda}(t) =  \pdv{l(\vecobs(t))}{\vecobs} - \blambda(t) \pdv{\obsop_{\btheta}(t, \vecobs(t),\vecobstaui{1}(t), \dots, \vecobstaui{\ndelays}(t))}{\vecobs} \\
      \displaystyle \qquad \quad - \sum_{i=1}^{\ndelays}{ \blambda(t + \tau_i) \pdv{\obsop_{\btheta}(t + \tau_i, \vecobs(t+\tau_i), \vecobstaui{1}(t+\tau_i), \dots, \vecobstaui{\ndelays}(t+\tau_i))}{\vecobstaui{i}}}  \\ 
      \blambda(t\geq t_T) = \bzero.
    \end{cases}\,$ 
    \STATE Compute $\displaystyle \pdv{L}{\btheta}$: \\ 
    \begin{align*}
    \begin{split} 
        \pdv{L}{\btheta}  = & - \!\!\!\int_{t_0}^T{ \!\!\!\!\!\blambda(t)\! \left(\pdv{\obsop_\btheta(t, \vecobs(t),\vecobstaui{1}(t), \dots, \vecobstaui{\ndelays}(t))}{\btheta} - \sum_{i=1}^{\ndelays}{\pdv{\obsop_\btheta(t, \vecobs(t), \vecobstaui{1}(t), \dots, \vecobstaui{\ndelays}(t))}{\vecobstaui{i}}} \right. } \\
        & \qquad \left. \pdv{\vecobs(t - \tau_i)}{\btheta} \right) \ddroit t
    \label{eq:grad_wtr_param_several_delays}
    \end{split}
    \end{align*}
    \STATE Update $\btheta$
\ENDFOR
\end{algorithmic}
\end{algorithm}


\subsection{On the importance of relevant delays} \label{App_MI}

Learning the delays $\tau_i$ within NDDEs is crucial for accurately modeling partially observed dynamics and, therefore, delays need to be adapted during training. To illustrate the impact of the relevant delays, we evaluate the information that past observations share with the current dynamics of the observable, and how this varies with the delay. To this end, let us consider a dynamical system evolving on a compact smooth manifold $\mathcal{S} \subset \mathbb{R}^\nstate$ described by a simple 2-delay dynamical system, assumed to be an attractor, and let us consider a $C^2$ observable function $\obs:\mathcal{S} \to \mathbb{R}$ 
\begin{eqnarray}
 \obs(t+\Delta t) & = & \cos(\obs(t-\tau_1)) \: \sin(\obs(t-\tau_2)) - \alpha \: \mathrm{sinc}(3 \, \obs(t-\tau_1)) + \alpha \: \cos(\obs(t-\tau_2)), \nonumber \\
\obs(t<0) & = & \psi(t), \label{eq:toydelay}
\end{eqnarray}
with $\alpha = 0.2$, $\tau_1 = p_1^\star \, \Delta t$, $\tau_2 = p_2^\star \, \Delta t$, $p_1^\star=250$, $p_2^\star=400$ and $\mathrm{sinc}(x) := \sin(x)/x$ for $x \ne 0$, $\mathrm{sinc}(0)$ := 1.

\medskip
The Takens' theorem \cite{takens} rigorously discusses the conditions under which a delayed vector of a real-valued observables $\left(\obs(\vecstate(t)), \obs(\vecstate(t-\tau)), \ldots, \obs(\vecstate(t-\ndelays \tau))\right)$, $\ndelays \in \mathbb{N}$, defines an embedding, a smooth diffeomorphism onto its image. It guarantees a topological equivalence between the original dynamical system and the one constructed from the memory of the observable. The dynamics of the system can then be reformulated on the set $\left(\obs(\vecstate(t)), \obs(\vecstate(t-\tau)), \ldots, \obs(\vecstate(t-\ndelays \tau))\right)$. However, Takens' theorem, later extended by \cite{sauer1991embedology}, provides a sufficient condition for reconstruction but does not specify how to choose the time delay $\tau$. From a mathematical viewpoint, the delay could be arbitrary, besides some specific values excluded by the theorem. In practice, its value can strongly condition a successful embedding. If too small, entries of the delay vector data are too similar; if too large, the entries tend to be completely uncorrelated and cannot be numerically linked to a consistent dynamical system \cite{kantz2003nonlinear}.

\begin{figure}[t]
\begin{center}
\includegraphics[width=0.6\textwidth]{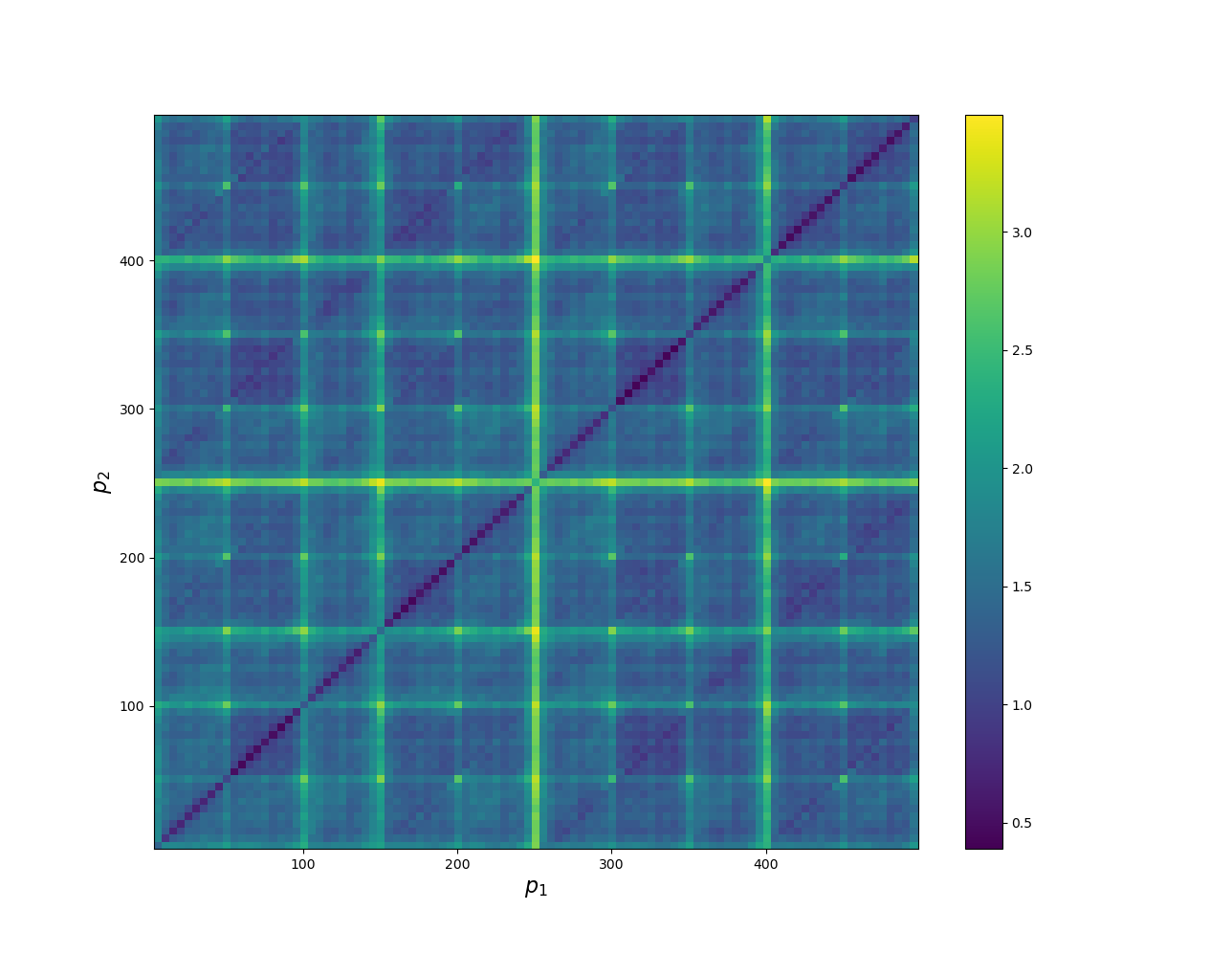}
\caption{$\left\{\tau_1 = p_1 \, \Delta t, \tau_2 = p_2 \, \Delta t\right\}$-map of Delayed Mutual Information, $I\left(\left(\obs(t-\tau_1),\obs(t-\tau_2)\right),\obs(t)\right)$. The maximum is exhibited at $(p_1, p_2)$ = $(250, 400)$ and $(400, 250)$, in agreement with $p_1^\star=250$, $p_2^\star=400$.}
\label{MItwotaus}
\end{center}
\end{figure}

\medskip
We here illustrate the impact of suitable delays in the relevance of the information available to inform the future evolution of the observable. The relevance of the delays $\left\{\tau_1, \tau_2\right\}$ for informing $\obs(t+\Delta t)$ is assessed in terms of the mutual information between $\left(\obs(t-\tau_1),\obs(t-\tau_2)\right)$ and $\obs(t)$, as suggested in \cite{fraser}. The mutual information $I(X,Y)$ between two random variables $X$ and $Y$ quantifies the amount of information observing a variable brings onto the other one. By considering the problem in Equation~\eqref{eq:toydelay}, we estimate the information shared between the current observations $\obs(t)$ and its past values $\obs(t-\tau_1)$, $\obs(t-\tau_2)$ and is shown in Fig.~\ref{MItwotaus} as a 2-D map in terms of $p_1$ and $p_2$. The map is symmetric, as expected since $I\left(\left(\obs(t-\tau_1),\obs(t-\tau_2)\right),\obs(t)\right) = I\left(\left(\obs(t-\tau_2),\obs(t-\tau_1)\right),\obs(t)\right)$. It can be seen that the amount of information shared between the current observation and a delay vector of the observable significantly varies with the delays and indeed reaches a maximum for $\tau_1=p_1^\star \, \Delta t$, $\tau_2=p_2^\star \, \Delta t$. The ability of the present Neural DDE method to learn the delays, in addition to the model $\obsop_\btheta$, is thus key to its performance and wide applicability.

\section{Experiments}\label{experiments}

We now demonstrate the present approach and its capability to jointly learn the delays in addition to a model of the dynamics. Several experiments are considered, ranging from synthetic low-dimensional dynamical systems to time-series predictions in an experimental configuration. In addition to accurately learning the dynamics, the relevance of being able to learn suitable delays is also demonstrated.


\subsection{Dynamical systems}\label{sec:dyn_system}

In the following the dynamical systems considered in this article are presented. All datasets are divided into training, validation, and test sets with proportions of 70\%, 10\%, and 20\%, respectively and the Dormand–Prince (Dopri5) solver was used for data generation and training \cite{dormand1980family}. Table~\ref{main_nb_delays} outlines the number of delays employed in NDDE for each experiment.

\subsubsection{Population dynamics model} As a first example, we consider the dynamics of a scalar-valued system used to model population dynamics in biology, \cite{pop_dyn1,pop_dyn2}. Such a system is formulated through the following DDE:
\begin{equation}
\dv{\state(t)}{t} = \state(t) \left( 1 - \state(t-\tau)\right), \qquad \state(t \leq 0) = \psi(t),
\end{equation}
where we integrate from $t \in [0, 10]$, $\tau = 1$, $\psi(t) = \state_0$, $\state_0$ is sampled from the uniform distribution $\mathcal{U}(2.0, 3.0)$ and $256$ trajectories were generated.


\subsubsection{Brusselator} A second experiment showcases how NDDEs can effectively model partially observed systems with past state values. We consider the 2-species Belousov-Zhabotinsky kinetic equation \cite{Belousov,Zhabotinskii} that can be modelled by the so-called Brusselator system:
\begin{equation}
\begin{aligned}
\begin{cases}
& \displaystyle \dv{\state_1(t)}{t} = A - B \state_1 - \state_1 + \state_1^2 \state_2, \\
& \displaystyle \dv{\state_2(t)}{t} = B \state_1 - \state_1^2 \state_2,
\end{cases}
\end{aligned}
\label{eqn:brusselator}
\end{equation}
with $\state_1(t)$ and $\state_2(t)$ the two species concentrations at a given time. We integrate in the time domain $t \in [0, 25]$. The initial condition $\state_1$ is sampled from the uniform distribution $\mathcal{U}(0, 2.0)$ while $\state_2 = 0.0$ and $1024$ trajectories were generated. We set ourselves in the partially observable case where we only have access to the dynamics of $\state_1$, \ie{}, $\obs(t) \equiv \state_1(t)$, and wish to reconstruct the whole dynamics.

\begin{figure}
    \centering
\begin{minipage}{0.45\textwidth}
    \includegraphics[width=\textwidth]{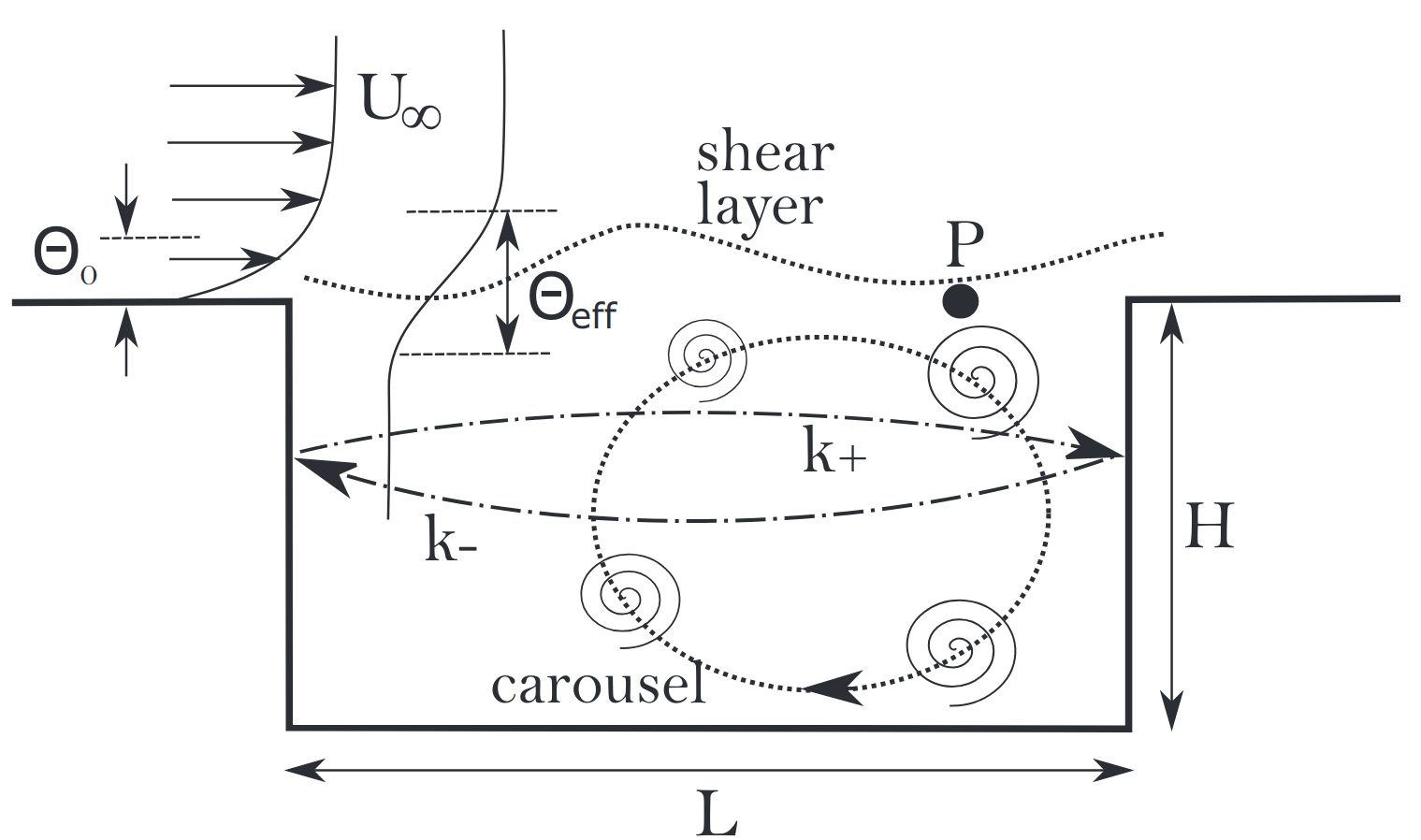}
    \captionof{figure}{Sketch of open cavity flow taken from \cite{tuerke2020nonlinear}. A data acquiring sensor is placed in P. The cavity has a length L and depth H. The incoming laminar boundary layer flow is characterized by the freestream velocity $U_{\infty}$ and the momentum thickness $\Theta_{0}$.}
    \label{fig:cavity_shear_flow}
\end{minipage}
\hfill
\begin{minipage}{0.45\textwidth}
\centering
    \begin{tabular}{lccc}
        \toprule
        & KS & Cavity & Brusselator \\
        \midrule
        NDDE & $5$ & $1$ & $2$ \\
        \bottomrule
    \end{tabular}
    \vspace{.75cm}
    \captionof{table}{Number of delays used in NDDE for each experiment.}
    \label{main_nb_delays}
\end{minipage}
\end{figure}


\subsubsection{Kuramoto-Sivashinsky (KS) system} This model was originally derived to describe the complex dynamics of flames in a combustion. The Kuramoto-Sivashinsky system is described as: 
\begin{equation}
\pdv{u}{t} + \pdv[2]{u}{x} + \pdv[4]{u}{x} + \frac{1}{2}\pdv{u^2}{x} = 0.\label{eqn:ks_equation}
\end{equation}
The system is integrated over the time domain $t \in [0,30]$ and its spatial domain $D_x = [0,22]$ is discretized into 128 points. A total of $2048$ trajectories were generated from the initial condition samples. To put ourselves in the partially observed setting, we choose to observe the solution in time at $\nobs$ locations uniformly spread across the spatial domain (we retain $\nobs=5$).


\subsubsection{Incompressible open cavity flow} As an experimental demonstrator, we consider the modeling based on time-series derived from wind tunnel experiments of an open cavity flow sketched in Figure~\ref{fig:cavity_shear_flow}. The facility and the details on the experimental setup are described in \cite{tuerke2020nonlinear}. Open cavity flows have attracted numerous research efforts in the last decades for the interesting dynamics at work. The flow is characterized by an impinging shear layer activating a centrifugal instability in a cavity. This interplay, reminiscent of the feedback acoustic mechanisms described in \cite{rossiter1964wind}, leads to a self-sustained oscillation. A broad range of dynamics is observed, ranging from limit cycles to toroidal and chaotic dynamics. The system is observed through a sensor in $P$ measuring the local pressure fluctuations, hence $\obs(t) \equiv P(t;\bx_{\mathrm P})$.
The data are obtained for a Reynolds number $Re=9190$ based on the length L of the cavity.


\subsection{Results}\label{results}

\begin{table}[b!]
\begin{center}
\begin{tabular}{lccccc} 
\toprule
& LSTM & NODE & ANODE & Latent ODE & NDDE \\
\midrule
Brusselator & $\mathbf{0.0051 \pm 0.0031}$ & $0.77 \pm 0.00080$ & $\mathbf{0.0050 \pm 0.0050}$ & $0.014 \pm 0.0076$ & $0.016 \pm 0.0076$ \\
KS & $0.77 \pm 0.041$ & $0.71 \pm 0.10$ & $0.55 \pm 0.027$ & $0.43 \pm 0.07$ & $\mathbf{0.28 \pm 0.024}$ \\
Cavity & $0.75 \pm 0.46$ & $0.96 \pm 0.0011$ & $0.65 \pm 0.0090$ & $0.25 \pm 0.14$ & $\mathbf{0.13 \pm 0.012}$ \\
\bottomrule
\end{tabular}
\caption{Model performance (MSE) over the test set on each experiments averaged over 5 runs.}
\label{tab:testloss}
\end{center}
\end{table}

We now assess the performance of the models with their ability to predict future measurements of a partially observed system. In this study, LSTM, NODE, ANODE, Latent ODE and NDDE were selected for comparison. Table~\ref{tab:testloss} displays the test MSE loss over each experiment. Appendix~\ref{hyperparam} goes in more details about each model's architecture and the training and testing procedure. Every model incorporates a form of "memory" into its architecture, with the exception of NODE. While LSTM and Latent ODE utilizes hidden units and ANODE employs its augmented state $\veca(t)$, NDDE leverages past observations such as $\vecobs(t-\tau)$. In all subsequent figures, the y-axis $\vecobs(t)$ represents our observables (introduced for each system in Section~\ref{sec:dyn_system}), defined as $\vecobs(t) = \obs(\vecstate(t))$.

\subsubsection{Population dynamics model} 
\begin{figure}[t]
    \centering
    \begin{minipage}{0.475\textwidth}
        \centering
        \includegraphics[width=1.1\textwidth,height=1.1\textheight, keepaspectratio]{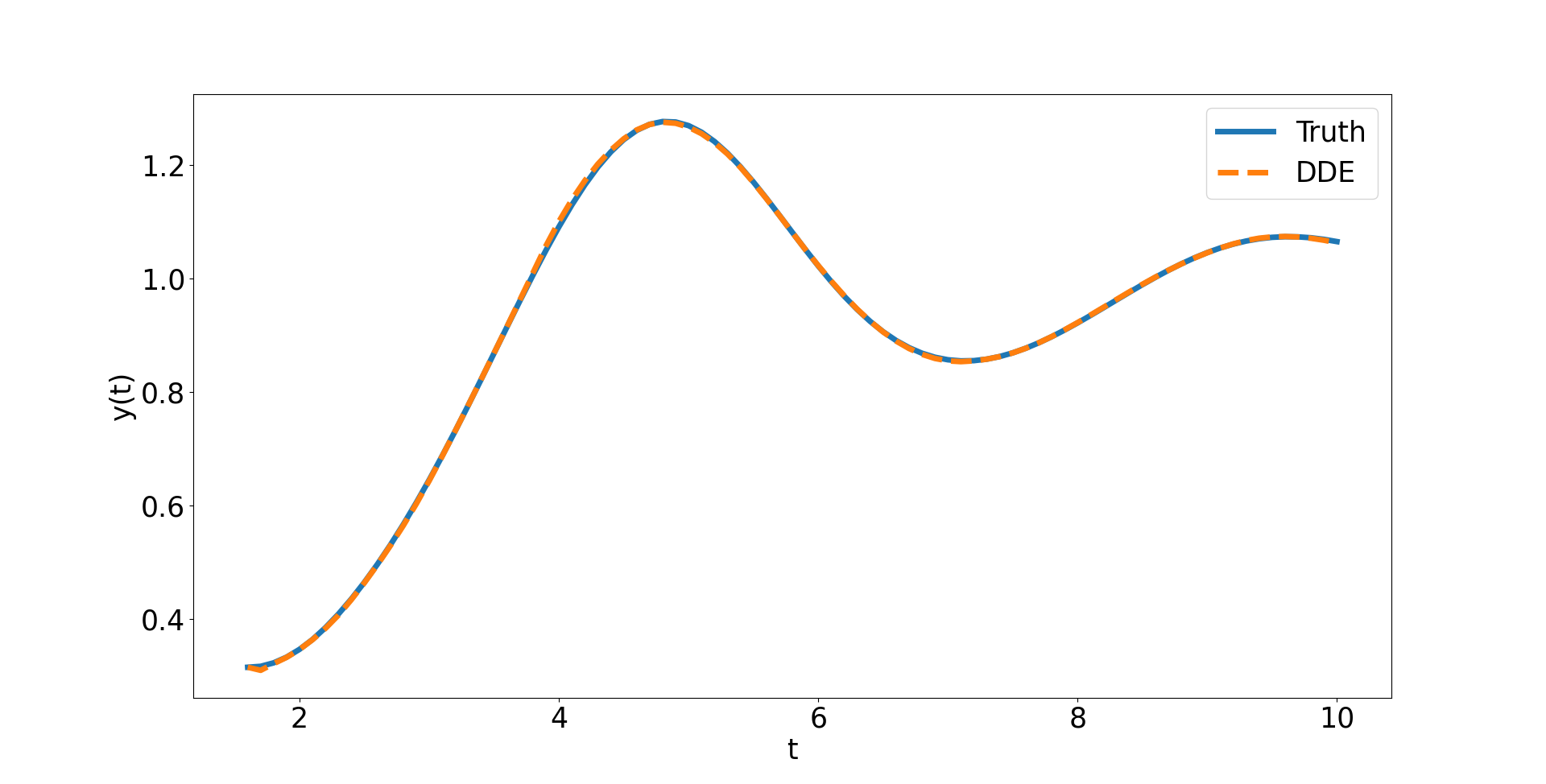}
        \caption{The prediction of the DDE model is seen to accurately match the true evolution.}
        \label{fig:testset_toydataset}
    \end{minipage}\hfill
    \begin{minipage}{0.475\textwidth}
        \centering
        \includegraphics[width=1.1\textwidth,height=1.1\textheight, keepaspectratio]{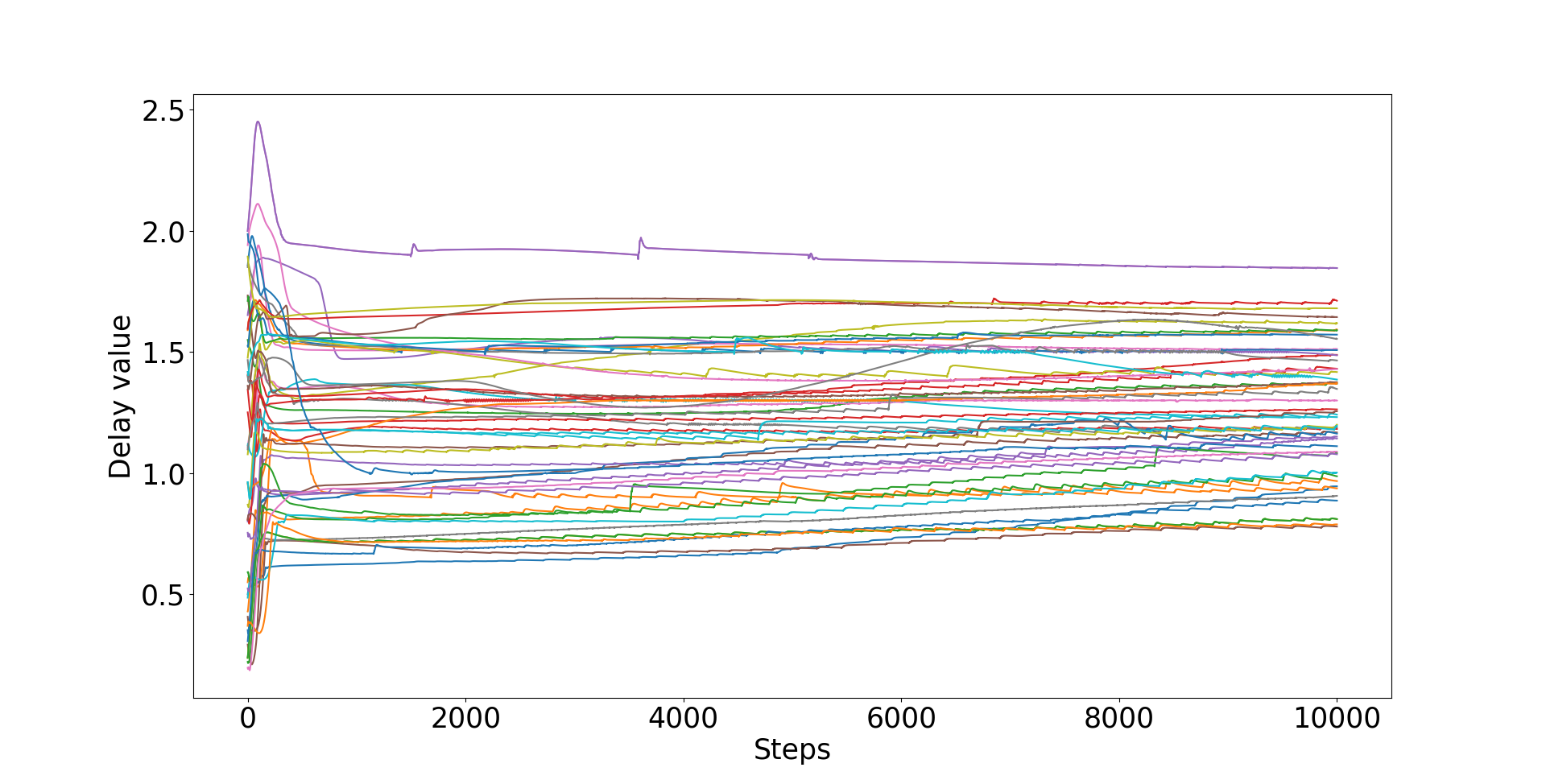} 
        \caption{Toy dataset delay evolution during training for several initial conditions.}
        \label{fig:delays}
    \end{minipage}
\end{figure}

Figure~\ref{fig:testset_toydataset} and~\ref{fig:delays} respectively depict the model's robust convergence to accurate dynamics and the evolution of the delay $\tau$ during training over many seeds, showcasing a consequence of Takens' theorem \cite{takens}. Using a delay-coordinate map, one can indeed construct a diffeomorphic shadow manifold $M^{\prime}$ from univariate observations of the original system in the generic sense. We here observe the whole state, so that $\obs(t) \equiv \state(t)$, and the delay coordinate map is then in terms of $(\obs(t), \obs(t-\tau))$, with $\tau >0$. The result of Figure~\ref{fig:delays} shows that the learned delay can lie between about 0.8 and 1.4. In classical approaches (see \cite{tan2023selecting}), the selected delay with Takens' theorem for State Space Reconstruction is typically chosen close to the minimum delayed mutual information of the time series. 
\begin{figure}[t]
\begin{minipage}[c]{0.475\linewidth}
\centering
\includegraphics[width=1.01\textwidth,height=1.01\textheight, keepaspectratio]{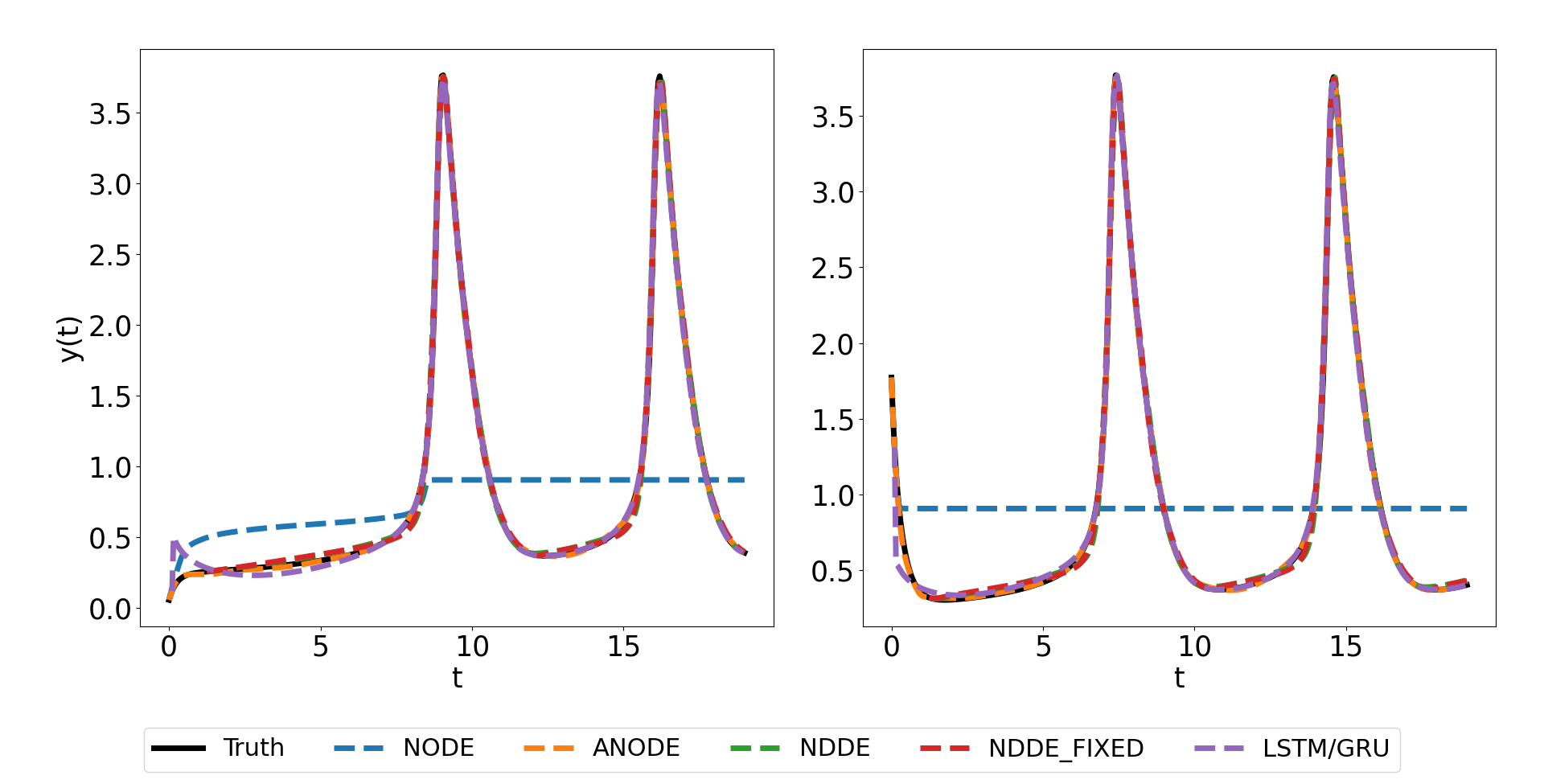}
\caption{Examples of prediction of the Brusselator dynamics in terms of $\obs(t)\equiv\state_1(t)$, for initial conditions sampled at random.}
\label{fig:brusselator_testsample1}
\end{minipage}
\hfill
\begin{minipage}[c]{0.475\linewidth}
\centering
\includegraphics[width=1.0\textwidth,height=1.0\textheight, keepaspectratio]{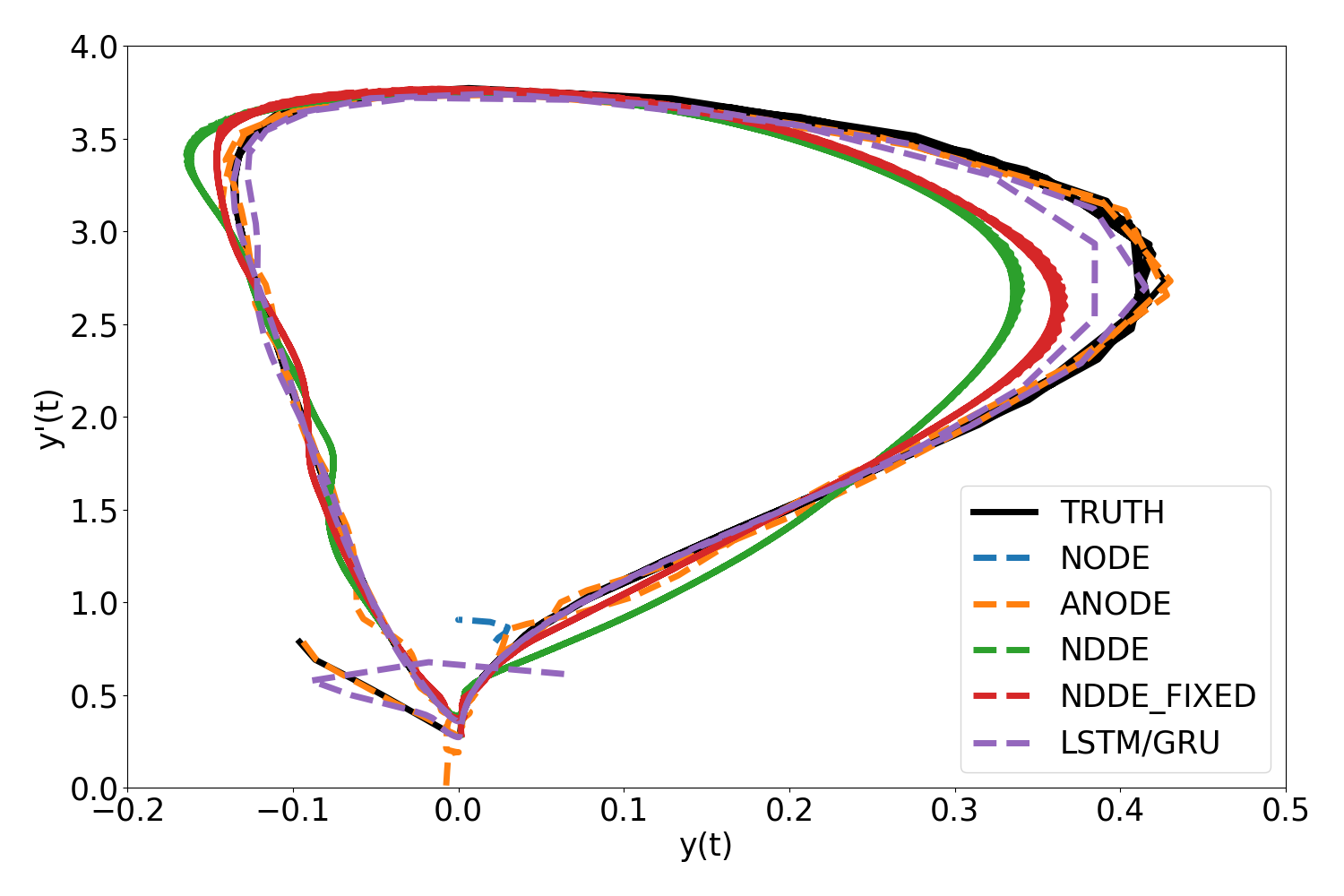}
\caption{Phase-portrait of $(\obs(t),\dot{\obs}(t))$. Long-term behavior of each trained model for the Brusselator system.}
\label{fig:brusselator_long_term}
\end{minipage}
\end{figure}

\subsubsection{Brusselator} In the case of this stiff and periodic dynamics, all models demonstrate satisfactory performance except NODE, which essentially predicts a mean trajectory thus highlighting the importance of incorporating memory terms for an accurate prediction. Remarkably, both LSTM and ANODE perform equally well, with NDDE and Latent ODE slightly trailing by a narrow margin as shown in Figure~\ref{fig:brusselator_testsample1}. In addition to evaluating the MSE loss performance, Figure~\ref{fig:brusselator_long_term} demonstrates the stability of each trained model on the Brusselator system over an extended period. After training within a specific time interval, we lengthened the integration period to five times the original training duration to assess the models' performance. It is observed that NDDE, along with LSTM, NODE and Latent ODE, are the only models that remain stable throughout this duration, with NDDE exhibiting the best performance over the extended horizon.

\subsubsection{KS system} 
\begin{figure}[t]
\centering
\includegraphics[width=0.75\textwidth, keepaspectratio]{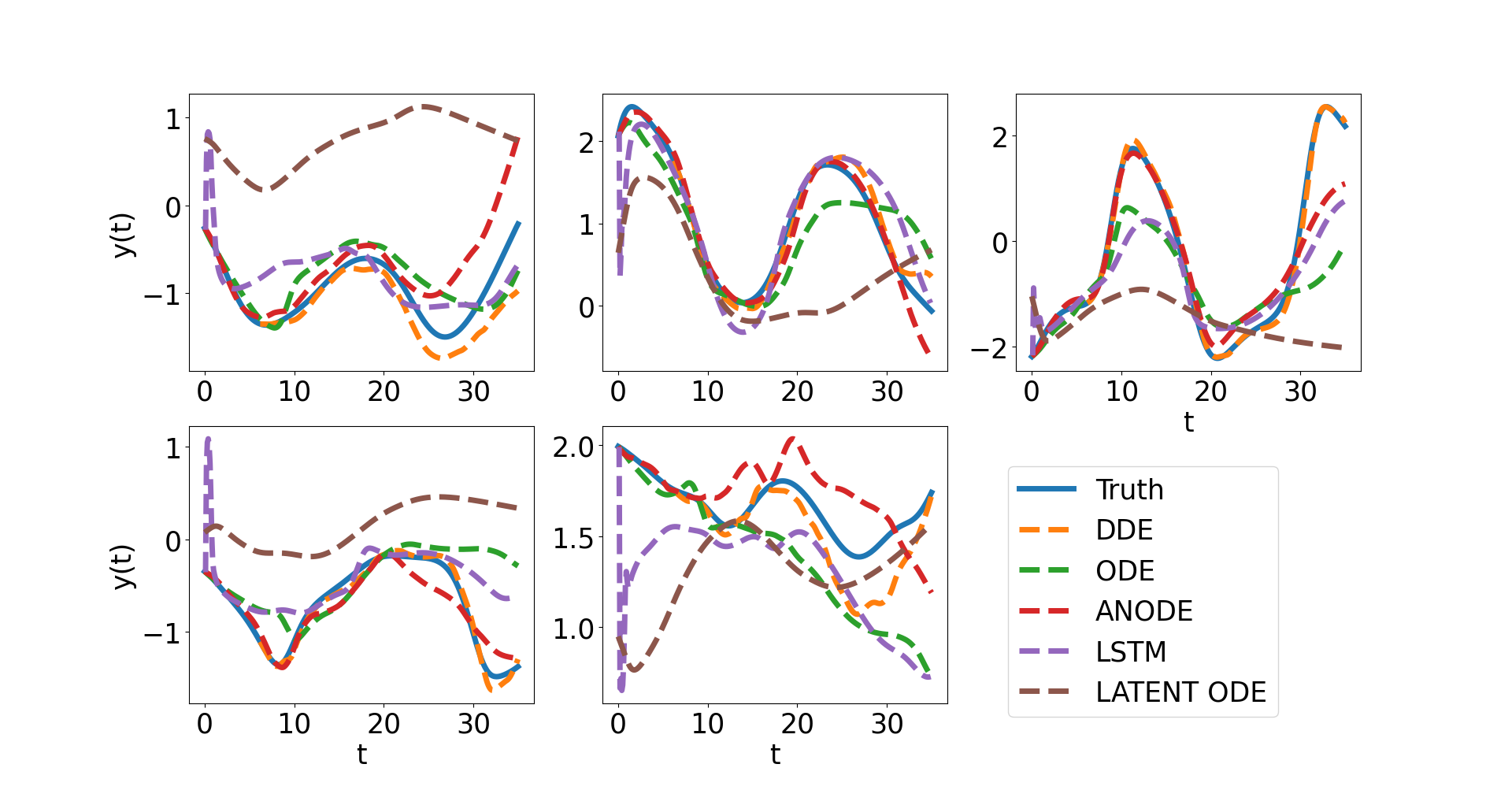}
\caption{Prediction of the KS system from a test sample for different models.}
\label{fig:ks_testsample}
\end{figure}

\begin{figure}[t]
\centering
\includegraphics[width=\textwidth,height=\textheight, keepaspectratio]{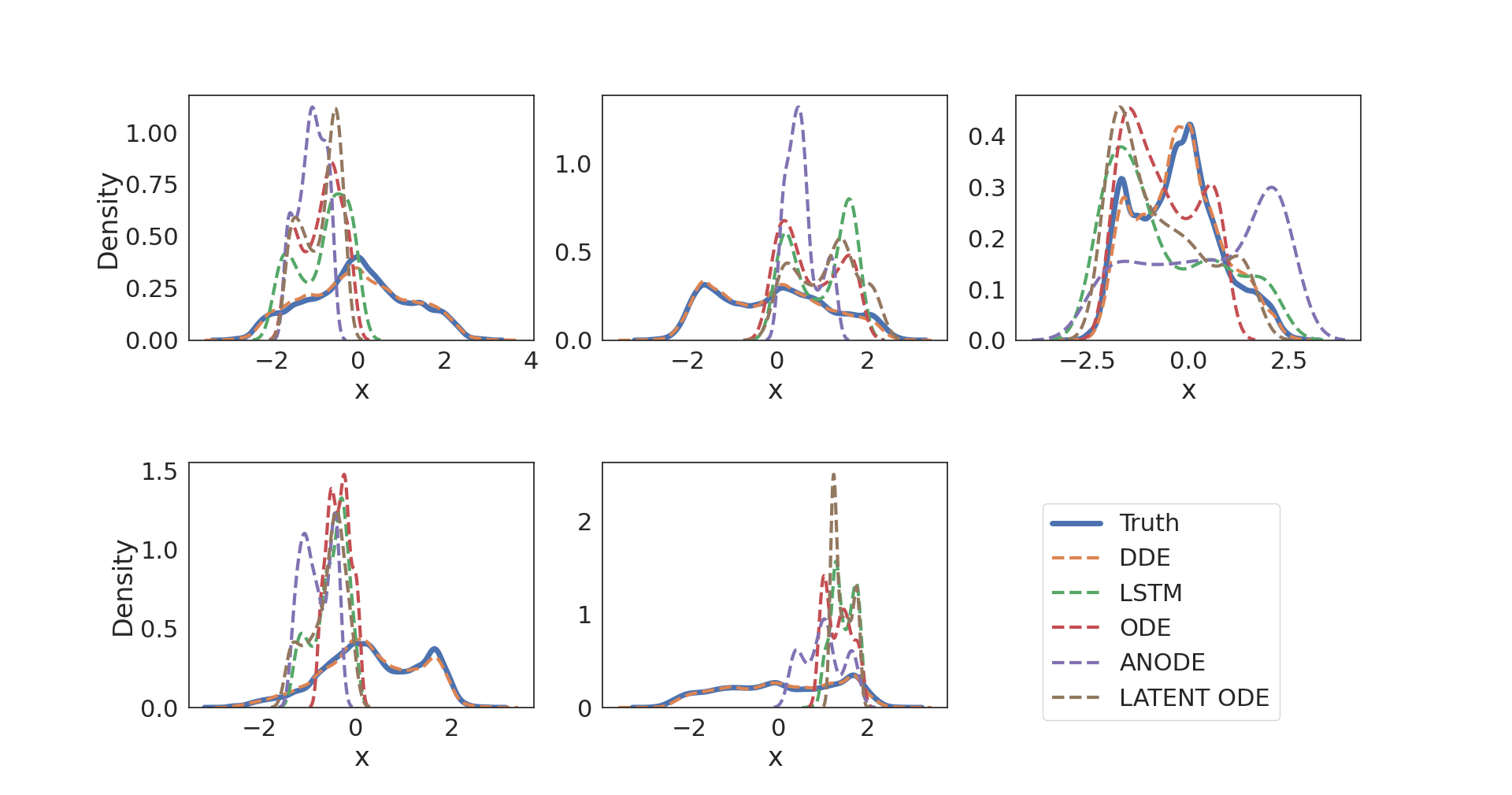}
\caption{Probability density functions of the predictions $y_i(t)$ over the time horizon  and across several initial conditions for different models. KS system.}
\label{fig:ks_density}
\end{figure}

This numerical experiment deals with a partially observed system in a chaotic regime, observed through the solution at $\nobs$ locations uniformly spread across the spatial domain. Figure~\ref{fig:ks_testsample} showcases random test samples from two different initial conditions, highlighting how NDDEs outperform other models struggling with the dynamics of the selected features. In a chaotic setting, the statistics of the dynamics are often more informative than the trajectory itself. 
We hence focus on the probability distribution of the prediction across a large time horizon for the different $\nobs$ components of the observables $\left\{\obs_i\right\}_{i=1}^\nobs$, \textit{cf.} Figure~\ref{fig:ks_density}. It is seen that NDDE again outperforms the other approaches, with predictions which are statistically closer to the ground truth. This is further supported by the evolution of the maximum Lyapunov exponent (MLE) of the resulting models. Table~\ref{tab:mle_ks} displays the MLE estimates for each model, showing that the Neural DDE with learnable delays closely aligns with the ground truth compared to other models.
\begin{table}[H]
    \centering
    \begin{tabular}{@{}lccccc@{}}
        \toprule
        & Ground Truth & NDDE   & NODE   & ANODE  & Latent ODE \\ \midrule
        $\blambda_{\rm max}$ & 0.129  & \textbf{0.128} & 0.097 & 0.120 & 0.035     \\ \bottomrule
    \end{tabular}
    \caption{Estimation of the maximum Lyapunov exponent $\blambda_{\rm max}$ for the KS system based on the generated trajectories from the test set for each model.}
    \label{tab:mle_ks}
\end{table}

\subsubsection{Cavity} 
\begin{figure}[t]
\begin{minipage}{0.5875\textwidth}
\centering
\includegraphics[width=1\textwidth,height=1\textheight, keepaspectratio]{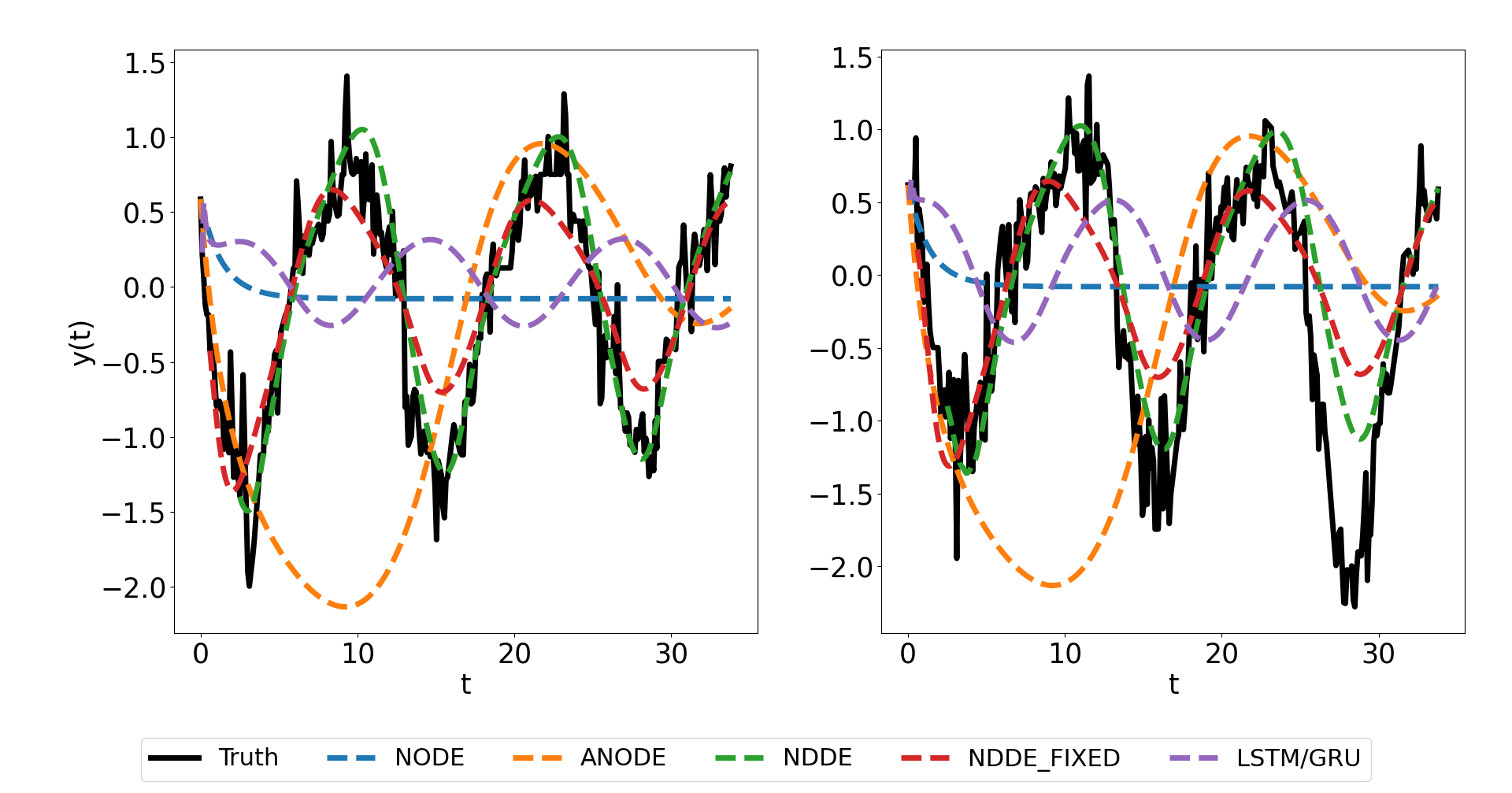}
\caption{Prediction of the cavity observables from different models for two test samples.}
\label{fig:cavity_test}
\end{minipage}\hfill
\begin{minipage}{0.3875\textwidth}
\centering
\includegraphics[width=1.0\textwidth,height=1.0\textheight, keepaspectratio]{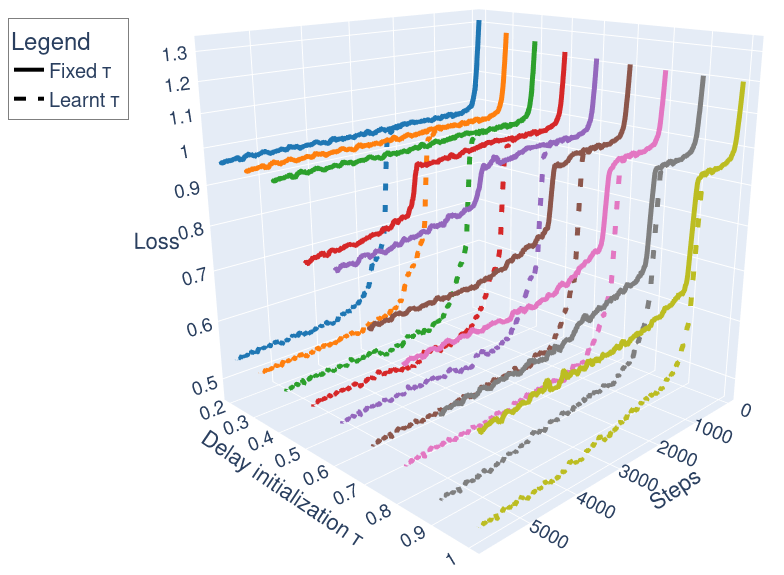}
\caption{Evolution of the MSE train loss of the NDDE model with constant (solid lines) and learnable delays (dashed lines) for different delay initialization values ranging from $0.2$ to $1.0$.}
\label{fig:cavity_flow_img}
\end{minipage}
\end{figure}

In this experimental fluid flow configuration, the NDDE formulation again outperforms other models as illustrated in Figure~\ref{fig:cavity_test} where the predictions are plotted for two test samples. The benefit of a time-delay model here comes from the fact that this experimental setup involves a large vortex within the cavity, effectively acting like a feedback loop and hence suitably described by governing equations involving delayed contributions. Latent ODE yields acceptable results compared to NODE which only generates the system's average trajectory, while LSTM and ANODE capture oscillations, albeit occasionally in conflicting phases. These experiments demonstrate that NDDE can effectively model trajectories even in the presence of noise in the data such as here with this experimental dataset.

Since the noise in the measurements is resulting from phenomena not related to the physical system at hand, such as for instance the electronic noise of the sensors, it is statistically independent from the true observables time-history. Learning to relate these past data with current measurements then quickly averages-out these spurious additional dimensions introduced by the noise as the amount of observations increases. More formally, the noise $\noise$ being assumed additive, low amplitude, and zero-mean, the expectation of the map $\obsop_{\btheta}$ writes

\hspace*{-0.2cm}\vbox{\begin{align}
 &\mathbb{E}\left[\obsop_{\btheta}(t, \vecobs(t) + \noise(t), \vecobs(t-\tau_1) + \noise(t-\tau_1), \dots, \vecobs(t-\tau_{\ndelays}) + \noise(t-\tau_{\ndelays}))\right] \nonumber
\\
& \qquad \qquad \approx \mathbb{E}\left[\obsop_{\btheta}(t, \vecobs(t), \vecobs(t-\tau_1), \dots, \vecobs(t-\tau_{\ndelays}))\right] + \mathbb{E}\left[\boldsymbol{\nabla} \obsop_{\btheta}(t, \vecobs(t), \vecobs(t-\tau_1), \dots, \vecobs(t-\tau_{\ndelays})) \cdot \noise\right], \nonumber \\
& \qquad \qquad = \mathbb{E}\left[\obsop_{\btheta}(t, \vecobs(t), \vecobs(t-\tau_1), \dots, \vecobs(t-\tau_{\ndelays}))\right] + \mathbb{E}\left[\boldsymbol{\nabla} \obsop_{\btheta}(t, \vecobs(t), \vecobs(t-\tau_1), \dots, \vecobs(t-\tau_{\ndelays}))\right] \cdot \mathbb{E}\left[\noise\right], \nonumber \\
& \qquad \qquad = \mathbb{E}\left[\obsop_{\btheta}(t, \vecobs(t), \vecobs(t-\tau_1), \dots, \vecobs(t-\tau_{\ndelays}))\right], \nonumber
\end{align}
}
so that the noise is effectively averaged-out provided the empirical mean approximates the expectation well enough.

To illustrate the benefit of learning the delays, instead of a priori setting them, the performance of NDDEs is here illustrated in both situations. A set of delays is considered, ranging from 0.2 to 1, and the Mean Squared Error (MSE) associated with each resulting models is monitored, with and without learning the delays, see Fig.~\ref{fig:cavity_flow_img}. The results show that learned delays consistently outperform fixed delays in leading to models associated with a significantly lower MSE.


\subsection{Closure modeling with the ROMs}\label{sec:dderom}

We now revisit the Kuramoto-Sivashinsky configuration from a closure modeling viewpoint. Conceptually, one here assumes an accurate model of the system to be known but the approximate time-dependent solution to be evaluated with a reduced-order model (ROM) Galerkin approach. Specifically, the solution $\vecstate(t)$ is approximated as $\displaystyle \vecstate(t) \approx \sum_{i=1}^{\nterms}{\varphi_i(\bx) \, a_i(t)}$, with $\left\{\varphi_i(\bx) \equiv \bvarphi_i\right\}_{i=1}^{\nterms}=:\Phi$ spanning a spatial basis, so that describing the solution boils down to deriving an evolution equation for the coefficients $\veca(t) = \left(a_1(t), \ldots, a_{\nterms}(t)\right)$. Because the approximation basis $\Phi$ is not spanning the whole state space whenever $\nterms < \mathrm{card}(\vecstate)$, the $\Phi$-projected model describing the dynamics of the coefficients $\left\{a_i(t)\right\}_i$, resulting from a Galerkin formulation, is not $\Phi$-invariant. To compensate for the loss of information leaking from the subspace spanned by $\Phi$, one can complement the reduced-order model with an additional, data-based, \emph{closure} term in the governing equations. Our experiment employs a Galerkin-projected ROM approach, with modes $\left\{\bvarphi_i\right\}_i$ derived from a Proper Orthogonal Decomposition (POD), with $\nterms$ = 4, 8, and 10 POD modes (low, medium, and high data regimes), respectively accounting for 58\%, 87\% and 94\% of the system's energy (\ie{}, variance). In practice, the more modes selected in our POD Galerkin ROM, the smaller the correction required by the closure term needs to be.  

\begin{figure}[t]
        \centering
        \includegraphics[width=0.8\textwidth]{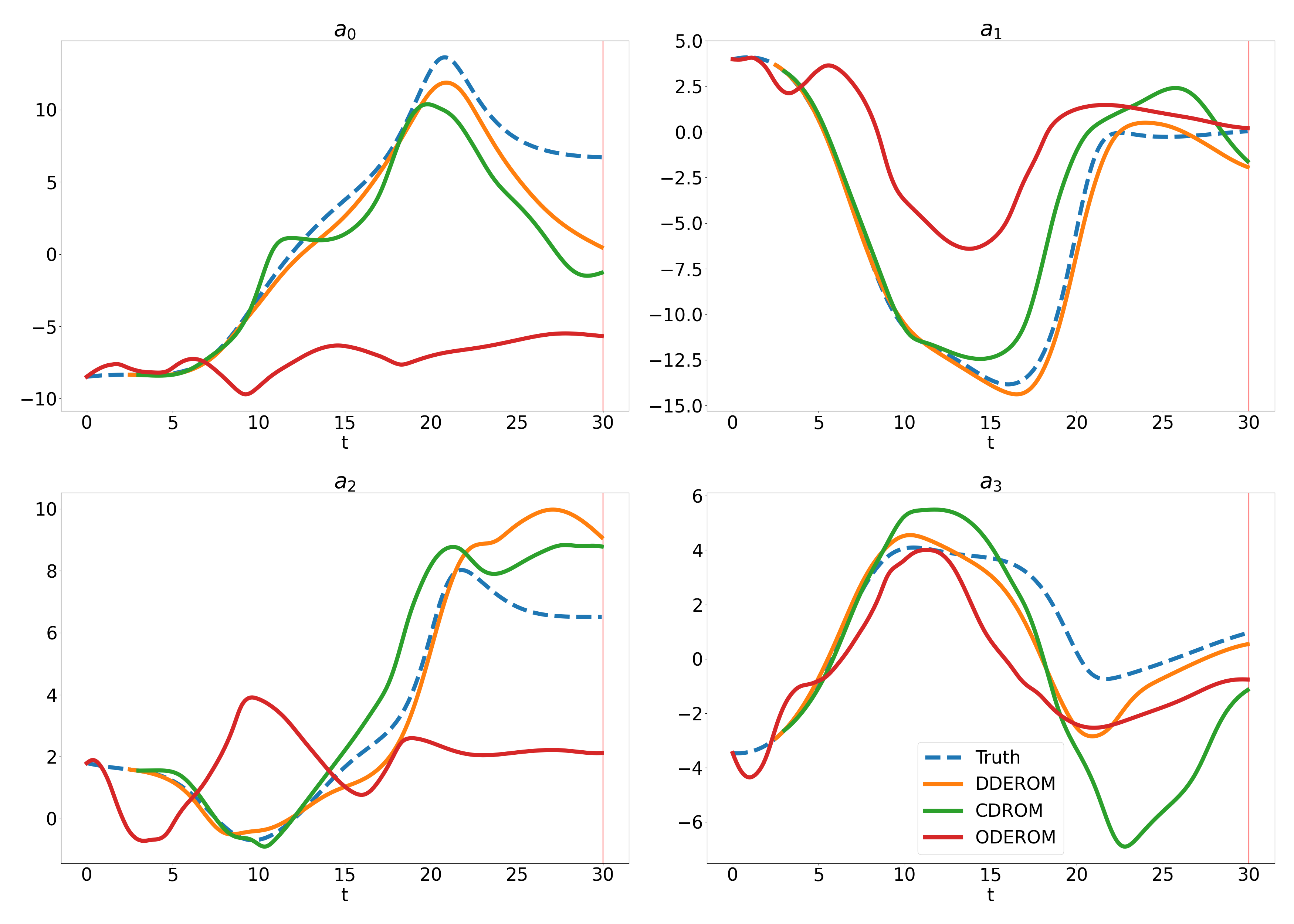}
        \caption{Examples of predictions of POD Galerkin ROM (4 modes).}
        \label{fig:ks_test_4_modes}
\end{figure}
\begin{figure}[t]
    \centering
    \includegraphics[width=\linewidth, keepaspectratio]{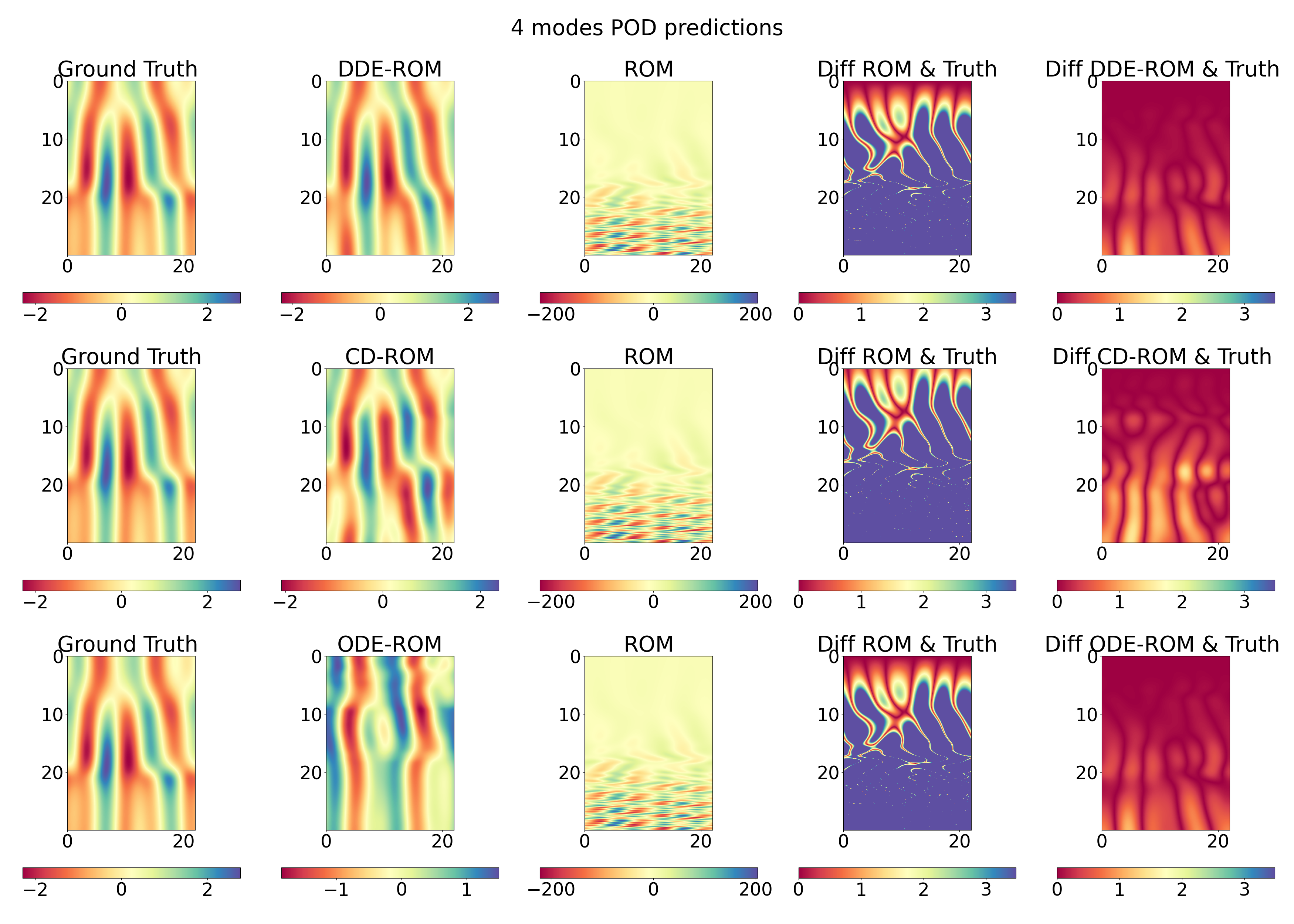}
    \caption{Examples of predictions of POD Galerkin ROM (4 modes) with different models from KS testset reconstructed in terms of full solution field $\state(\bx,t)$ (First three columns). The two rightmost columns show the absolute value of the reconstruction error associated with each models.}
    \label{fig:ks_test_reconstructed_4_modes}
\end{figure}

\medskip
We now consider different models for representing the closure term and compare their performance in terms of the test MSE loss in Table \ref{tab:table_mse_pod}. It is seen that the DDE closure term with learnable delays consistently outperforms the ODE closure term across all data regimes (low, medium, and high). This performance difference becomes especially pronounced in the low data regime of the 4-mode POD Galerkin ROM, indicating that the Neural ODE closures are not well-suited for such scenarios. Figure \ref{fig:ks_test_4_modes}, along with their respective representations in the original state space in Figure \ref{fig:ks_test_reconstructed_4_modes}, showcase the prediction from two samples of initial conditions from the test dataset for the 4-mode POD Galerkin ROM. As the number of modes increases, the discrepancy between ODE and DDE closures decreases quantitatively, as seen in Table \ref{tab:table_mse_pod}. This trend is expected since the ROM then captures most of the system's information. By studying these specific ROMs, we can identify where ODE closure terms fall short and how incorporating past states with the DDE closure term can address ODE's deficiencies in low data scenarios. Our approach is compared with the closure model CD-ROM \cite{Menier_JCP} which considers a continuous embedding of past information in the form of an exponentially decaying dynamics auxiliary term $\bz(t)$ mimicking a memory:
\begin{equation}
    \bz(t) = \int_{t_0}^t{e^{(t-\tau) \, \Lambda} \, \vecobs(\tau) \, \ddroit \tau}
\end{equation}
with $\Lambda$ a diagonal matrix whose elements are associated with the relevant time-scales of the system under consideration; see \cite{Menier_JCP} for details. Interestingly, compared to a memoryless Markovian model such as NODE, it only improves the quality of the prediction only in the low-data regime and considering a higher dimensional memory variable $\bz(t)$ does not improve the learning process in high-data regimes.

Several numbers of learnable delays were considered for the NDDE approach, ranging from 1 to 3. For this system, the results were essentially unaffected by the number of delays, showing that a single delay is here enough to recover most of the necessary information for a good prediction.

\begin{table}[H]
    \centering
    \begin{tabular}{llll}
    \toprule
    Model & 4 modes & 8 modes & 10 modes \\ 
    \midrule
    CD-ROM & $7.94 \pm 0.53$ & $0.43 \pm 0.24$ & $0.099 \pm 0.019$ \\
    ODE-ROM & $13.61 \pm 0.34$ & $0.44 \pm 0.044$ & $0.084 \pm 0.0028$ \\
    DDE-ROM & $\mathbf{3.39 \pm 0.034}$ & $\mathbf{0.18 \pm 0.07}$ & $\mathbf{0.067 \pm 0.0078}$ \\
    \bottomrule
    \end{tabular}
    \caption{Summary of model performance metrics, \ie, the test MSE loss. CD-ROM, ODE, and DDE closure models are compared across the 4, 8, and 10 POD modes settings.}
    \label{tab:table_mse_pod}
\end{table}


\section{Conclusion}\label{sec:conclusion}

We introduced Neural Delay Differential Equations (NDDEs) as a principled and data-efficient framework for modeling partially observed dynamical systems. This approach is motivated on one hand by the Mori-Zwanzig framework first developed in the statistical physics community for accounting for the effect of unobservable quantities onto observed ones; and on the second hand by the Takens' theorem.  Relying on these frameworks, we leverage the link between unresolved dynamics and explicit time-delay representations via NDDEs, which allow for a continuous-time, physically interpretable, representation of non-Markovian dynamics. In particular, we show that the memory term can be approximated with a finite number of learnable delays without loss of generality under smooth dynamics, providing a compact yet expressive formulation for modeling systems with memory.

\medskip
Methodologically, we proposed an adjoint-based training procedure for NDDEs with learnable delays, enabling efficient end-to-end optimization over both model parameters and delay variables. The accompanying open-source implementation, \texttt{torchdde}, provides an accessible and reproducible platform for future research and applications.

\medskip
Numerical and experimental validations are discussed to assess the efficacy of NDDEs across different settings, including synthetic, chaotic, and real-world noisy data such as the Kuramoto-Sivashinsky equation, and experimental data from cavity-flow configurations. The performed experiments revealed two key insights: first, the pivotal role of a form of memory in accurately capturing dynamics; second, it was demonstrated that LSTMs' and Latent ODEs' hidden latent states or ANODEs' latent variables sometimes come short to achieve optimal performance, emphasizing the efficacy of delayed terms as an efficient dynamics memory mechanism. In fact, across all scenarios, NDDEs consistently outperformed or matched state-of-the-art continuous-depth and recurrent models while maintaining a smaller parameter footprint and improved interpretability. Finally, we revisited the Kuramoto-Sivashinsky configuration using NDDEs as a closure model of a reduced-order model based on Proper Orthogonal Decomposition modes. With respect of current state-of-art closure models, the present approach again exhibits superior performance. 

\medskip
In conclusion, Neural Delay Differential Equations provide a theoretically grounded and computationally efficient alternative to traditional recurrent and latent-state neural models for complex non-Markovian dynamical systems. Future research will aim at further generalizing the framework; while overestimating the number of delays does not affect the final performance, rigorously determining the optimal number of delays to consider in NDDE is a challenge. A potentially useful step toward a more principled and meaningful estimation might be achieved with promoting the delayed mutual information between past data and prediction of the future observations \cite{fraser}.


\paragraph{Acknowledgments}
We thank Dr. Emmanuel Menier for helpful discussions.

\paragraph{Funding Statement}
This work has been funded by the French National Agency for Research under project \# ANR-20-CE23-0025-01.

\paragraph{Competing Interests}
None

\paragraph{Data Availability Statement}
The codes associated to the examples discussed in the article can be found at the following github repositories: \href{https://github.com/thibmonsel/learnable_delays}{https://github.com/thibmonsel/learnable\_delays} and \href{https://github.com/thibmonsel/DDEROM}{https://github.com/thibmonsel/DDEROM}. The \texttt{torchdde} library for learning jointly the delays and the DDE's dynamics in these models is made opensource at \href{https://github.com/thibmonsel/torchdde}{https://github.com/thibmonsel/torchdde}.

\paragraph{Ethical Standards}
The research meets all ethical guidelines, including adherence to the legal requirements of the study country.

\paragraph{Author Contributions}
Conceptualization: T.M., O.S., G.C., L.M. Methodology: T.M., O.S., G.C., L.M. Data curation: T.M. Data visualisation: T.M. Writing original draft: T.M., O.S., L.M.  All authors approved the final submitted draft.


\newpage

\appendix

\section{Neural IDE and Neural DDE Benchmark}\label{app:idevsdde}

Firstly, let us compare both Neural IDE and Neural DDE analytically where any function $\stateop_{\btheta}$ here denotes a parameterized network:

\begin{align}
   \frac{\ddroit \vecobs}{\ddroit t} & = M_{\btheta}(\vecobs(t)) + \int_0^t{K_{\btheta}(\vecobs(t-s),s)   \ddroit s}, \label{ap:ide} \\
   \frac{\ddroit \vecobs}{\ddroit t} & = \obsop_{\btheta_1}(\vecobs(t)) + \obsop_{\btheta_2}(t, \vecobs(t), \vecobs(t -\tau_1),\dots, \vecobs(t-\tau_{\ndelays})). \label{ap:dde}
\end{align}

The second term on the right hand side of Equation~\eqref{ap:ide} is much more computationally involved than the second term on the right hand side (RHS) of Equation~\eqref{ap:dde}. Indeed, Equation~\eqref{ap:dde} only needs 2 function evaluations to evaluate its right hand side while, on the other hand, the number of function evaluation required to integrate Equation~\eqref{ap:ide} increases as $t$ grows in order to get a correct estimate of the integral term. For a more theoretical examination of computational complexity, one can refer to Appendix A.6 of \cite{monsel2024time}.

\subsection{Computation Time}

\begin{figure}[H]
\centering
\includegraphics[width=.675\textwidth,height=.675\textheight, keepaspectratio]{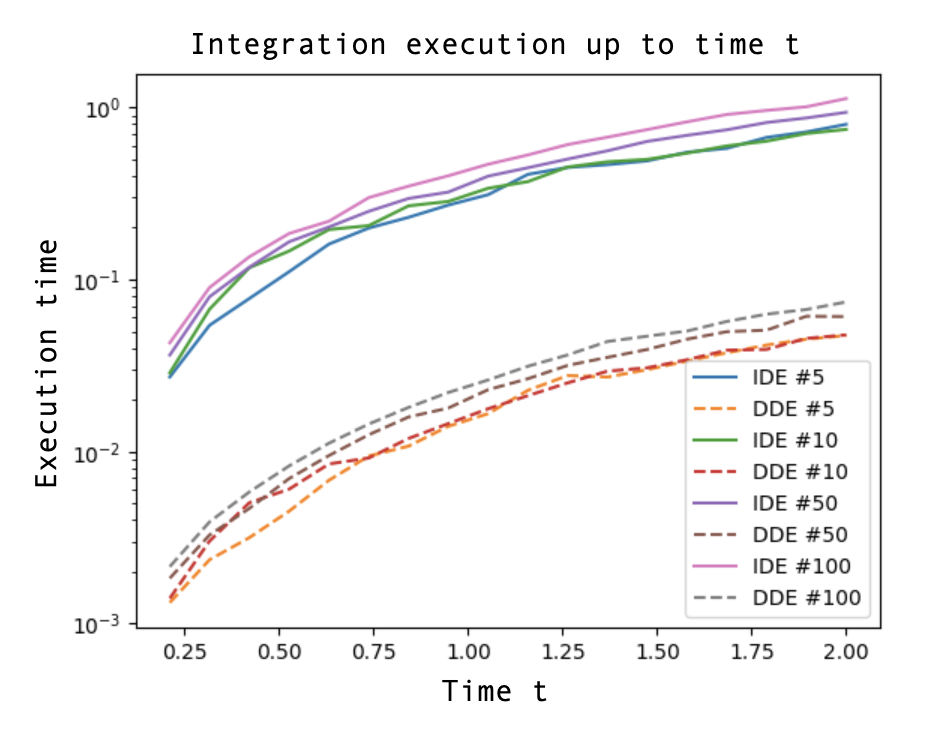}
\caption{Computation time of the forward pass (averaged over 5 runs) as a function of the size of the horizon for different numbers of observation (5, 10, 50, 100).}
\label{fig:idevsdde}
\end{figure}

Figure~\ref{fig:idevsdde} compares the computation time of a forward pass between a Neural IDE and Neural DDE, as considered above of similar size (roughly $500$ parameters), with the following setup:

\begin{itemize}
    \item Runge-Kutta 4 (RK4) solver with a timestep of $dt=0.1$,
    \item Batch size of $128$,
    \item Neural DDE uses 5 delays,
    \item Different numbers of observations are considered: $\left\{5,10,50,100\right\}$,
    \item The upper integration bound is varied from $t=0.2$ to $t=2.0$.
\end{itemize}

This benchmark clearly shows how expensive Neural IDE is compared to NDDE. NDDEs integration is at least an order of magnitude faster. Note that in Figure~\ref{fig:idevsdde}, the notation "\#i" refers to the number of observables $\obs_i(\vecstate,t)$.

\subsection{\texttt{torchdde}'s memory and time benchmark \label{torchddebench}}

In this subsection, we provide time and memory benchmarks on some of \texttt{torchdde}'s solvers. In order to compare both training methods of optimize-then-discretize (\ie, the adjoint method) and discretize-then-optimize (\ie, regular backpropagation), we present the Brusselator's experiment time duration and memory usage for various solvers during training (with a batch size of 1024) in Tables~\ref{clocktime} and~\ref{gpu_memory}, respectively. The adjoint method is slower (by a small factor) and requires less memory than the regular backpropagation. These results are consistent with NODE's examination of the adjoint method and conventional backpropagation tradeoffs, \cite{chen2018neural}.

\begin{table}[H]
\centering
\begin{minipage}{0.45\textwidth}
\centering
\begin{tabular}{lcc}
\toprule
  & Adjoint & Backpropagation \\
\midrule
RK4    & $4.8 \pm 0.23$  & $1.89 \pm 0.09$ \\
RK2    & $2.4 \pm 0.005$ & $0.90 \pm 0.005$ \\
Euler  & $1.5 \pm 0.01$  & $0.47 \pm 0.003$ \\
\bottomrule
\end{tabular}
\caption{Clock time (s) per batch.}
\label{clocktime}
\end{minipage}%
\hfill
\begin{minipage}{0.45\textwidth}
\centering
\begin{tabular}{lcc}
\toprule
        & Adjoint         & Backpropagation \\
\midrule
RK4     & $2.2 \pm 18$    & $2.87 \pm 4$     \\
RK2     & $2.15 \pm 20$   & $2.48 \pm 3$     \\
Euler   & $2.09 \pm 15$   & $2.264 \pm 9$    \\
\bottomrule
\end{tabular}
\caption{GPU consumption (Gb $\pm$ Mb) per batch.}
\label{gpu_memory}
\end{minipage}
\end{table}

\begin{figure}[H]
\centering
\begin{minipage}{0.45\textwidth}
    \centering
    \includegraphics[width=\textwidth,height=0.5\textheight,keepaspectratio]{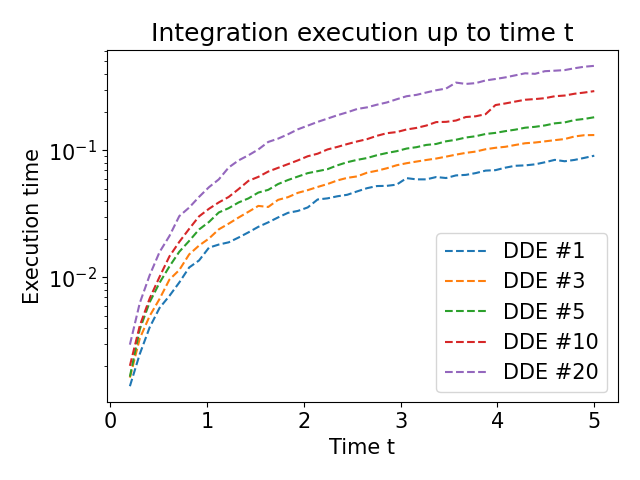}
    \caption{Time duration of forward pass averaged over 5 runs.}
    \label{fig:dde_nb_delays_time}
\end{minipage}%
\hfill
\begin{minipage}{0.45\textwidth}
    \centering
    \includegraphics[width=\textwidth,height=0.5\textheight,keepaspectratio]{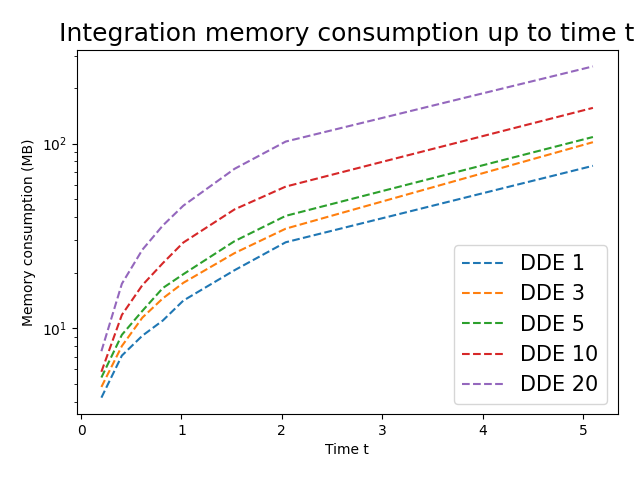}
    \caption{Memory consumption of forward pass averaged over 5 runs.}
    \label{fig:dde_nb_delays_mem}
\end{minipage}
\end{figure}

Figure~\ref{fig:dde_nb_delays_time} and~\ref{fig:dde_nb_delays_mem} compares respectively time and memory consumption for a forward of a Neural DDE with varying number of delays, each having approximately 28,000 parameters. We use the following setup: 

\begin{itemize}
    \item Use a RK4 solver with a timestep of $dt=0.1$,
    \item Use a batch size of $128$,
    \item Neural DDE uses $\left\{1, 3, 5, 10, 20\right\}$ delays,
    \item The number of observations is $\nobs = 100$,
    \item The upper integration bound is varied from $t=0.2$ to $t=5.0$,
    \item The notation ``\#i'' in the figure refers to the number of delays used in the Neural DDE.

\end{itemize}
        
\section{Derivation of Proposition~\ref{th:adjoint_dde}}\label{app:proof_adjoint}

We want to solve the constrained optimization problem introduced in Section~\ref{Sec_learning_delays}. The delays are part of the learning and are hence a subset of the parameter vector $\btheta$ to be optimized. The derivation is first presented for a single delay before being extended to the multiple delay situation below.
For convenience, we denote $\vecobstau(t)$ the third set of variables that $\obsop_\btheta(t,\vecobs(t),\vecobs(t-\tau))$ depends on, and with a dot the time derivative $\dot{\vecz}(t)$ of any quantity $\vecz(t)$. To recall, we consider the following minimization problem:

\begin{equation}
 \begin{aligned}
    & \minLM_{\btheta} \:  L(\vecobs),\\
    \mathrm{s.t.} \qquad  & L(\vecobs) = \int_{t_0}^T{l(\vecobs(t)) \ddroit t},\\
    & \dv{\vecobs(t)}{t} - \obsop_\btheta(t, \vecobs(t),\vecobs(t - \tau)) = \bzero,\\
    & \vecobs(t \leq t_0) = \bpsi(t).
\end{aligned}   
\end{equation}

We consider the following Lagrangian $\Lagr$:

\begin{equation}
   \Lagr := L(\vecobs) + \int_{t_0}^T{\blambda(t) \left( \dv{\vecobs(t)}{t} - \obsop_\btheta(t, \vecobs(t),\vecobstau(t)) \right) \ddroit t}.
   \label{Eq_def_Lagr}
\end{equation}

We use $\vecobs(t-\tau) = \vecobstau(t)$ wherever convenient. At optimality, the state equation is satisfied ($\vecobs$ is solution to the associated DDE), implying that $\displaystyle \pdv{\Lagr}{\btheta} = \pdv{L}{\btheta}$.

Integration by parts of the expression of the Lagrangian (Eq.~\eqref{Eq_def_Lagr}) yields:
\begin{equation}
    \Lagr = \int_{t_0}^T{l(\vecobs(t)) \ddroit t + \left[\blambda(t) \vecobs(t) \right]_{t_0}^T  - \int_{t_0}^T \left(  \dot{\blambda}(t) \vecobs(t) + \blambda(t) f_\btheta(t, \vecobs(t),\vecobstau(t)) \right) \ddroit t}.
\end{equation}

Stationarity of the Lagrangian at optimality yields $\displaystyle \frac{\partial \Lagr}{\partial \vecobs} = \bzero$ and expresses as:
\begin{align}
    \bzero & = \int_{t_0}^{T}{\left(\pdv{l(\vecobs(t))}{\vecobs} - \dv{\blambda(t)}{t} - \blambda(t) \pdv{f_\btheta(t,\vecobs(t),\vecobstau(t))}{\vecobs} - \blambda(t) \pdv{f_\btheta(t,\vecobs(t),\vecobstau(t))}{\vecobstau} \pdv{\vecobs(t-\tau)}{\vecobs} \right) \ddroit t} + \pdv{\left[\blambda(t) \vecobs(t) \right]_{t_0}^T}{\vecobs}
    \label{Eq_adjoint_v1}
\end{align}

With the change of variable $t \xrightarrow{} t - \tau$, the fourth term of the integrand rewrites as: 
\begin{align}
-\int_{t_0}^{T}{\blambda(t) \pdv{f_\btheta(t,\vecobs(t),\vecobstau(t))}{\vecobstau} \pdv{\vecobs(t-\tau)}{\vecobs} \ddroit t} & = - \int_{t_0-\tau}^{T-\tau}{\blambda(t+\tau) \pdv{f_\btheta(t+\tau,\vecobs(t+\tau),\vecobstau(t+\tau))}{\vecobstau} \pdv{\vecobs(t)}{\vecobs} \ddroit t}, \nonumber \\
& = - \int_{t_0}^{T}{\blambda(t+\tau) \pdv{f_\btheta(t+\tau,\vecobs(t+\tau),\vecobstau(t+\tau))}{\vecobstau} \ddroit t}
\end{align}
where we have set $\blambda(t) = \bzero$ whenever $t\ge T$ and recognized that $\vecobs(t \leq t_0) \equiv \bpsi(t)$ with $\displaystyle \pdv{\bpsi(t)}{\vecobs} = \bzero$.

The boundary terms $\displaystyle \pdv{\left[\blambda(t) \vecobs(t) \right]_{t_0}^T}{\vecobs}$ in Eq.~\eqref{Eq_adjoint_v1} then vanish and the equation is satisfied whenever the integrand is identically zero. Hence the following adjoint equation, to be integrated backwards in time:
\begin{align}
    \begin{split}
    & \dv{\blambda(t)}{t} = \pdv{l(\vecobs(t))}{\vecobs} - \blambda(t)  \pdv{f_\btheta(t,\vecobs(t),\vecobstau(t))}{\vecobs} - \blambda(t+\tau) \pdv{f_\btheta(t+\tau,\vecobs(t+\tau),\vecobstau(t+\tau))}{\vecobstau},\\
     &\blambda(t\geq T) = \bzero.
    \end{split}
    \label{Eq_adjoint_final}
\end{align}\label{eq:proof_adj_dyn_App}

Now taking the derivative of the Lagrangian with respect to $\btheta$ expresses:

\begin{align}\label{eq:grad1}
\begin{split}
    \pdv{\Lagr}{\btheta}  =  &\int_{t_0}^{T}{\pdv{l(\vecobs(t))}{\vecobs}\pdv{\vecobs(t)}{\btheta} \ddroit t} + \blambda(T) \pdv{\vecobs(T)}{\btheta} - \blambda(t_0) \pdv{\vecobs(t_0)}{\btheta} \\
    & + \int_{t_0}^T{ - \dot{\blambda}(t)  \pdv{\vecobs(t)}{\btheta} \ddroit t} + \int_{t_0}^T{ -\blambda(t) \pdv{f_\btheta(t, \vecobs(t),\vecobstau(t))}{\btheta} \ddroit t}  \\
    & + \int_{t_0}^{T}{ -\blambda(t) \pdv{f_\btheta(t, \vecobs(t),\vecobstau(t))}{\vecobs} \pdv{\vecobs(t)}{\btheta} \ddroit t} \\
    & + \int_{t_0}^{T}{ -\blambda(t) \pdv{f_\btheta(t, \vecobs(t),\vecobstau(t))}{\vecobstau} \pdv{\vecobs(t - \tau)}{\btheta} \ddroit t}.
\end{split}
\end{align}

Since $\displaystyle \pdv{\vecobs}{\btheta} \equiv \bzero$ and owing to the adjoint dynamics in Equation~\eqref{Eq_adjoint_final}, it finally comes

\begin{align}\label{eq:proof_grad_loss2}
    \begin{split}
        \pdv{L}{\btheta} = & - \int_{t_0}^T{ \blambda(t) \left(\pdv{f_\btheta(t, \vecobs(t),\vecobstau(t))}{\btheta} + \pdv{f_\btheta(t, \vecobs(t),\vecobstau(t))}{\vecobstau} \pdv{\vecobs(t - \tau)}{\btheta} \right) \ddroit t}.
    \end{split}
\end{align}

\subsection*{Case for multiple constant delays}

In the case of multiple delays $\tau_i$, $i \in {1, \dots, \ndelays}$, the equations in Theorem~\ref{th:adjoint_dde} are extended as follows. For each $i$, we define $\vecobstaui{i}(t) \equiv \vecobs(t-\tau_i)$. 

The second term $\displaystyle \blambda(t) \pdv{f_\btheta(t,\vecobs(t),\vecobstau(t))}{\vecobs}$ in the adjoint dynamics (Eq.~\eqref{Eq_adjoint_final}) is replaced by: 

\begin{align}
	\blambda(t) \pdv{f_\btheta(t,\vecobs(t),\vecobstaui{1}(t), \dots, \vecobstaui{\ndelays}(t))}{\vecobs}
\end{align}

and the last term $\displaystyle \blambda(t+\tau) \pdv{f_\btheta(t+\tau,\vecobs(t+\tau),\vecobstau(t+\tau))}{\vecobstau} $ by the following: 

\begin{align}
    \sum_{i=1}^{{\ndelays}}{\blambda(t + \tau_i) \pdv{f_\btheta(t+\tau_i, \vecobs(t+\tau_i), \vecobstaui{1}(t+\tau_i), \dots, \vecobstaui{\ndelays}(t+\tau_i))}{\vecobstaui{i}}}.
\end{align}

The first term  $\displaystyle \pdv{f_\btheta(t, \vecobs(t),\vecobstau(t))}{\btheta}$ in the gradient's Equation~\eqref{eq:proof_grad_loss2} becomes

\begin{align}
    \pdv{f_\btheta(t, \vecobs(t),\vecobstaui{1}(t), \dots, \vecobstaui{\ndelays}(t))}{\btheta}
\end{align}

and $\displaystyle \pdv{f_\btheta(t, \vecobs(t),\vecobstau(t))}{\vecobstau} \pdv{\vecobs(t - \tau)}{\btheta}$ is now: 
\begin{align}
    \sum_{i=1}^{{\ndelays}}{\pdv{f_\btheta(t, \vecobs(t), \vecobstaui{1}(t), \dots, \vecobstaui{\ndelays}(t))}{\vecobstaui{i}} \pdv{\vecobs(t - \tau_i)}{\btheta}}.
\end{align}


\section{Derivation of Proposition~\ref{prop:dde}}
\label{app:takens}

Let us start by stating Takens' theorem as expressed by \cite{noakes1991takens,takens}:

\begin{theorem}[Takens' embedding theorem] \label{th:takens} Let $\mathcal{M}$ be a compact space. There is an open dense subset $\mathcal{D}$ of $\mathrm{Diff}(\mathcal{M}) \times C^2(\mathcal{M}, \mathbb{R})$, with $\mathrm{Diff}(\mathcal{M})$ the diffeomorphism group of $\mathcal{M}$, with the property that the Takens map $$\embedd: \mathcal{M} \to \mathbb{R}^{2m+1}$$
given by $\embedd(\vecstate) = (\vecobs(\vecstate), \vecobs(\phi(\vecstate)), \vecobs(\phi\circ\phi(\vecstate)), \ldots, \vecobs(\phi^{2m}(\vecstate)))$  is an embedding of $C^2$ manifolds, where $(\phi, \vecobs) \in \mathcal{D}$. 
\end{theorem}

Here, $\phi$ stands for the operator that advances the dynamical system by a time step $\tau$, \ie, that sends $\vecstate(t)$ to $\vecstate(t+\tau)$, and $\vecobs$ is the observable operator, that sends a full state $\vecstate(t)$ to actual observables $\vecobs(\vecstate(t)) =: \vecobs(t)$.
Variants of this Theorem, \eg, \cite{sauer1991embedology}, include the consideration of any set of different delays $\tau_i$ instead of uniformly spaces ones.
The representation
$\embedd(\vecstate(t)) = (\vecobs(\vecstate(t)), \vecobs(\vecstate(t-\tau)), \vecobs(\vecstate(t-2\tau)), \ldots \vecobs(\vecstate(t-2m\tau))$
then becomes
$\embedd(\vecstate(t)) = (\vecobs(\vecstate(t)), \vecobs(\vecstate(t-\tau_1)), \vecobs(\vecstate(t-\tau_2)), \ldots \vecobs(\vecstate(t-\tau_{2m}))$.
In the proof of Takens' theorem, $m$ is the intrinsic dimension of the dynamical system, \ie, the one of the manifold $\mathcal{M}$.

Now, given the full state $\vecstate$ that follows the dynamics:

\begin{equation}
\begin{aligned}
\dv{\vecstate}{t} = \stateop(\vecstate), \qquad \vecstate(0)=\vecstate_0,
\end{aligned}
\end{equation}

we use the chain rule on the observable $\vecobs$: 
\begin{equation}
\begin{aligned}
&\dv{\vecobs}{t} = \vecobs^{\prime}(\vecstate(t)) \, \stateop(\vecstate).
\end{aligned}
\end{equation}

By applying the inverse of the delay coordinate map $\embedd^{-1}$ from Theorem~\ref{th:takens}, which is invertible from its image as it is an embedding, we show that the dynamics of $\vecobs$ possess a DDE structure:

\begin{equation}
\begin{split}
 \dv{\vecobs}{t} = (\vecobs^{\prime} \times \stateop) \circ \embedd^{-1}(\vecobs(t), \vecobs(t-\tau_1), \dots, \vecobs(t-\tau_n)).
\end{split}
\label{eqn:takens_equation}
\end{equation}

The proof is completed by choosing $\obsop = ( \vecobs^{\prime} \times \stateop) \circ \embedd^{-1} - M$ where $M$ is obtained by the Mori-Zwanzig formalism (Eq.~(\ref{eqn:mz_no_noise})).

Note that to be able to apply Takens' theorem, we needed the step-forward operator $\phi$ to be a diffeomorphism, \ie, the flow of the dynamical system to be smooth and smoothly invertible. Also, we used the differentiability of the observables $\vecobs$ to express $\vecobs^{\prime}$.

\section{Training hyperparameters}\label{hyperparam}

Our training approach incorporates a progressive strategy considered to be curriculum learning strategy, \cite{soviany2022curriculumlearningsurvey}. We begin by feeding the models shorter trajectory segments and gradually increase their length when the so-called \emph{patience} hyperparameter is exceeded. This process continues until we reach the desired trajectory length. Each time the trajectory length is increased, we reset the patience hyperparameter to 0. This \textit{patience} is then incremented by 1 if the validation loss fails to decrease, and reset to 0 if the validation loss improves. This method aligns with the principles of curriculum learning, a technique that involves training machine learning models in a structured order, typically progressing from simpler to more complex examples. In our case, this translates into moving from shorter to longer trajectories. This approach aims to improve the learning process and the resulting model performance. Table~\ref{patience_length} displays the patience parameter and the length of the trajectory considered initially. Table~\ref{numberparams} refers to the number of training parameters of each model. The loss function used across all experiments is the MSE loss, and we employ the Adam optimizer with a weight decay of $10^{-7}$. Table~\ref{learningrates} provides the initial and final learning rates ($\lrate_i$, $\lrate_f$) for each experiment, which are associated with the scheduler. The scheduler is a StepLR scheduler with a gamma factor ($\gamma = \exp(\frac{\log(\lrate_f \slash \lrate_i)}{N})$, with $N$ the trajectory's length). The scheduler adjusts the learning rate as the trajectory length increases, allowing training to start with the initial learning rate $\lrate_i$ and gradually decreases to the final learning rate $\lrate_f$. All continuous-time models (NODE, ANODE, Latent ODE and NDDE) used Runge-Kutta for numerical integration. Table~\ref{numberparams_mlp} shows the width and depth of the Multi Layer Perceptrons (MLPs) for NODE, ANODE, and NDDE across all experiments. Additionally, we provide the hidden size and number of layers for the LSTM model in Table~\ref{hiddenlayersparams_lstm}. Finally, Table~\ref{configurationparams_latent_ode} summarizes the Latent ODE hyperparameters, where the vector field $f_\btheta$ (defined in the Introduction) is an MLP with the width and depth specified in the second and third columns respectively, the size of the latent variable $z_0$ in the last column, and the RNN's hidden size in the fourth column.
If some models have less parameters compared to others it is that we found that they provided better results with less. ANODE's augmented state dimension matches that of the number of delays used by NDDE displayed in Table~\ref{main_nb_delays}.

\begin{table}[H]
\centering
\begin{tabular}{lccc}
\toprule
 & KS & Cavity & Brusselator  \\
\midrule
Length Start & 15\% & 50\% & 25\%  \\  
Patience & 40 & 50 & 20 \\
\bottomrule
\end{tabular}
\caption{How long is the trajectory chunks given at first and the patience used for each experiment. \label{patience_length}}
\end{table}

\begin{table}[H]
\centering
\begin{tabular}{@{}lccccc@{}}  
\toprule
& LSTM & NODE & ANODE & Latent ODE & NDDE \\
\midrule
Brusselator & $1764$ & $3265$ & $3395$ & $3666$ & $3331$ \\
KS & $18130$ & $9029$ & $11609$ & $8118$ & $19343$ \\
Cavity & $2234$ & $2209$ & $2274$ & $3642$ & $2242$ \\
\bottomrule
\end{tabular}
\caption{Number of parameters for each experiment. \label{numberparams}}
\end{table}

\begin{table}[H]
\centering
\begin{tabular}{@{}lcc@{}}  
\toprule
& \multicolumn{2}{c}{NODE/ANODE/NDDE} \\
\cmidrule(lr){2-3}
& Width & Depth \\
\midrule
Brusselator & 32 & 4 \\
KS & 64 & 3 \\
Cavity & 32 & 3 \\
Shallow Water & 32 & 3 \\
\bottomrule
\end{tabular}
\caption{MLP width and depth for each experiment. \label{numberparams_mlp}}
\end{table}

\begin{table}[H]
\centering
\begin{tabular}{@{}lcc@{}}  
\toprule
Experiment & Hidden Size & Number of Layers \\
\midrule
Brusselator & 5 & 10 \\
KS & 25 & 5 \\
Shallow Water & 6 & 10 \\
Cavity & 7 & 7 \\
\bottomrule
\end{tabular}
\caption{Hidden size and number of layers for each experiment for LSTM model. \label{hiddenlayersparams_lstm}}
\end{table}

\begin{table}[H]
\centering
\begin{tabular}{@{}lcccc@{}}  
\toprule
Experiment & Width Size & Depth & Hidden Size & Latent Size \\
\midrule
Brusselator & 16 & 3 & 16 & 16 \\
KS & 32 & 3 & 16 & 16 \\
Cavity & 16 & 3 & 8 & 8 \\
Shallow Water & 32 & 3 & 8 & 8 \\
\bottomrule
\end{tabular}
\caption{Configuration parameters for each experiment for Latent ODE. \label{configurationparams_latent_ode}}
\end{table}

\begin{table}[H]
\centering
\begin{tabular}{@{}lll@{}}  
\toprule
Experiment &  $\lrate_i$ & $\lrate_f$ \\
\midrule
Brusselator & 0.001 & 0.0001 \\
Cavity & 0.005 & 0.00005 \\
KS & 0.01 & 0.0001 \\
Shallow Water & 0.001 & 0.00001 \\
\bottomrule
\end{tabular}
\caption{Initial and final learning rates for each experiment. \label{learningrates}}
\end{table}


\bibliographystyle{alpha}
\bibliography{biblio}

\end{document}